\documentclass[final,5p,times,twocolumn,numbers]{elsarticle}

\usepackage{amssymb}
\usepackage{lipsum}
\usepackage{amsmath}
\usepackage{array}
\usepackage{tabularx}
\usepackage{supertabular}
\usepackage{pdflscape}
\usepackage{rotating}
\usepackage{threeparttable}
\usepackage{booktabs}
\usepackage{longtable}
\usepackage{url} 
\usepackage{tikz}
\usepackage{hyperref}
\usepackage{array}
\usepackage{float} 
\usepackage{pifont}   
\usepackage{amssymb}  
\usepackage{float}
\usepackage{multirow}

\usetikzlibrary{mindmap,shadows,trees}
\usetikzlibrary{mindmap,trees} 

\journal{Elsevier}
\usepackage{lineno}

\begin{document}
\begin{frontmatter}


\title{RF-DETR Object Detection vs YOLOv12 : A Study of Transformer-based and CNN-based Architectures for Single-Class and Multi-Class Greenfruit Detection in Complex Orchard Environments Under Label Ambiguity}

\author[1]{Ranjan Sapkota\corref{cor1}}
\author[2]{Rahul Harsha Cheppally}
\author[2]{Ajay Sharda}
\author[1]{Manoj Karkee \corref{cor1}}

\affiliation[1]{organization={Biological \& Environmental Engineering},
            addressline={Cornell University},
            city={Ithaca},
            postcode={14850},
            state={NY},
            country={USA}}
            
\affiliation[2]{organization={Department of Biological and Agricultural Engineering},
            addressline={Kansas State University},
            city={Manhattan},
            postcode={66502},
            state={KS},
            country={USA}}

\cortext[cor1]{Corresponding Authors: Manoj Karkee and Ranjan Sapkota}
\ead{mk2684@cornell.edu, rs2672@cornell.edu}
\begin{abstract}
This study presents a comprehensive comparison between RF-DETR object detection and YOLOv12 object detection models for greenfruit recognition in complex orchard environments characterized by label ambiguity, occlusion, and background camouflage. A custom dataset was developed featuring both single-class (greenfruit) and multi-class (occluded and non-occluded greenfruits) annotations to assess model performance under real-world conditions. The RF-DETR object detection model, leveraging a DINOv2 backbone with deformable attention mechanisms, excelled in global context modeling, which proved particularly effective for identifying partially occluded or visually ambiguous greenfruits. Conversely, the YOLOv12 model employed CNN-based attention mechanisms to enhance local feature extraction, optimizing it for computational efficiency and edge deployment suitability. In the single-class detection scenarios, RF-DETR achieved the highest mean Average Precision (mAP@50) of 0.9464, showcasing its robust capability to accurately localize greenfruits within cluttered scenes. Despite YOLOv12N achieving the highest mAP@50:95 of 0.7620, RF-DETR object detection model consistently outperformed in managing complex spatial scenarios. In multi-class detection, RF-DETR again led with an mAP@50 of 0.8298, demonstrating its effectiveness in distinguishing between occluded and non-occluded fruits, whereas YOLOv12L topped the mAP@50:95 metric with 0.6622, indicating superior classification under detailed occlusion conditions. The analysis of model training dynamics revealed RF-DETR's rapid convergence, particularly in single-class scenarios where it plateaued at fewer than 10 epochs, underscoring the efficiency and adaptability of transformer-based architectures to dynamic visual data. These results confirm RF-DETR’s suitability for accuracy-critical agricultural tasks, while YOLOv12 remains ideal for speed-sensitive deployments.
\end{abstract}





\end{frontmatter}

\textbf{Keywords:} Object Detection, \sep RF-DETR object detection model (Roboflow Detection Transformer), \sep YOLOv12 object detection model, \sep You Only Look Once, \sep Transformers, \sep CNNs, \sep Greenfruit Detection, \sep Deep Learning , \sep Machine-Vision 

\section{Introduction}
\label{introduction}
\begin{figure}[ht!]
\centering
\includegraphics[width=0.99\linewidth]{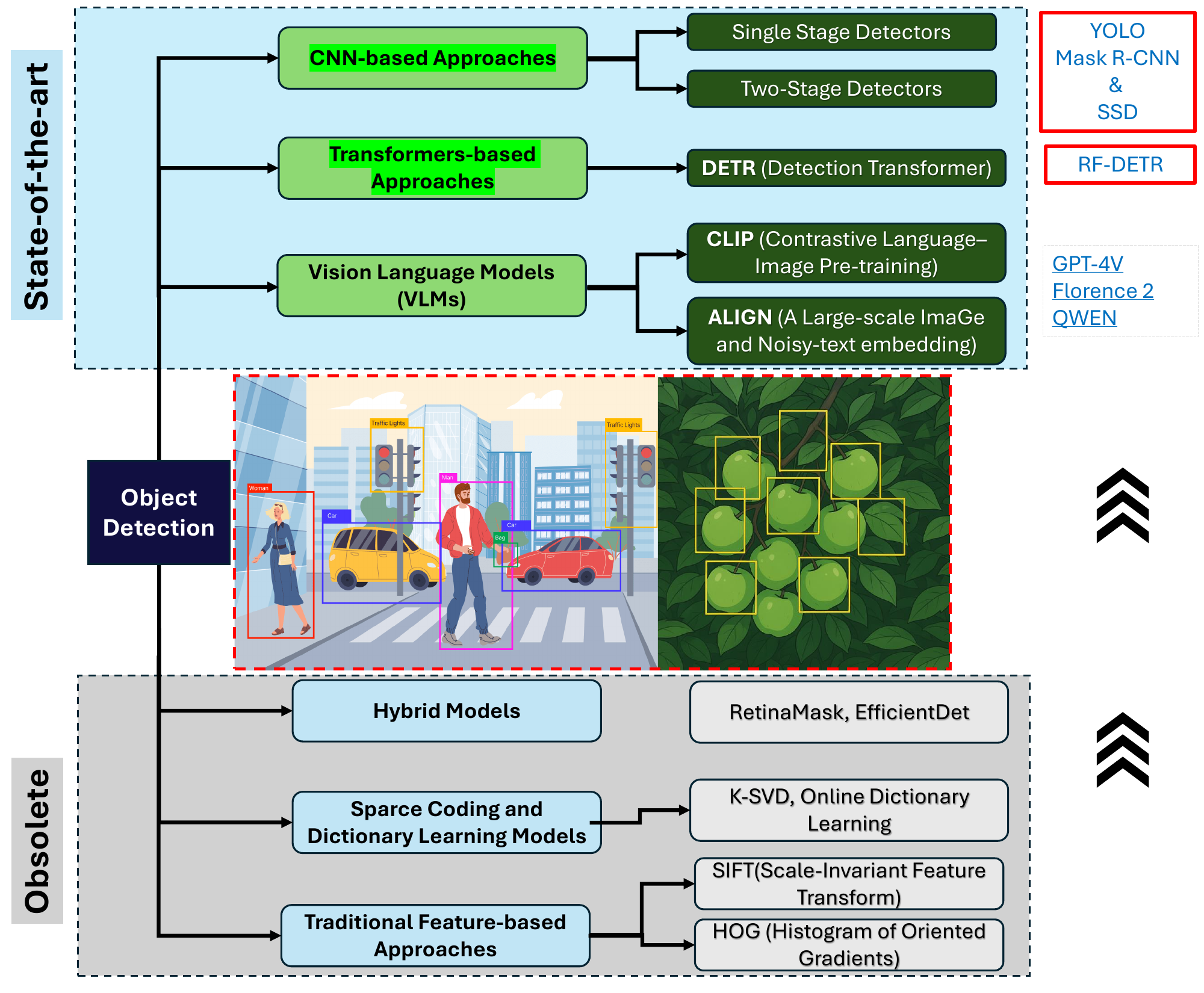}
\caption{\textbf{Classification of object detection methodologies: Top features state-of-the-art CNN-based and Transformer-based methods, widely adopted; Vision Language Models are emerging. Also includes Hybrid, Sparse Coding, and Traditional Feature-based approaches.}}
\label{fig:img1}
\end{figure}

As illustrated in Figure \ref{fig:img1}, the field of object detection over the past decade, propelled by breakthroughs in deep learning, have shifted from basic pattern recognition to sophisticated systems capable of complex image understanding. The object detection approaches can be branched into six primary methodologies as illustrated in Figure \ref{fig:img1}, each with unique strengths and applications in various domains of technology and automation. This evolution is vital for overcoming common visual recognition challenges in fields requiring high precision and adaptability, such as autonomous driving \cite{masmoudi2019object, hnewa2020object}, healthcare \cite{elakkiya2021cervical}, security surveillance \cite{mishra2016study}, and notably in agriculture \cite{badgujar2024agricultural, sa2016deepfruits}, where accurate and efficient object detection supports advancements like automated field monitoring \cite{singh2024iot} and robotic harvesting \cite{yang2024development}. 

The six approaches depicted in Figure \ref{fig:img1} are Convolutional Neural Networks (CNNs) \cite{gu2018recent} , Transformer-based models \cite{nimma2024intelpvt, liu2025transformer}, Vision Language Model-based approaches \cite{zang2025contextual, fu2025llmdet}, Hybrid models such as RetinaMask and EfficientDet \cite{fu2019retinamask, tan2020efficientdet}, Sparse Coding and Dictionary Learning models, and Traditional Feature-based approaches. Among them, CNNs, including the YOLO (You Only Look Once) series \cite{redmon2016you, sapkota2024comprehensive} and R-CNN (Region-based CNN) family such as Mask R-CNN \cite{he2017mask}, have become staples in practical deployments due to their proficient handling of spatial hierarchies. Transformer-based models like DETR (Detection Transformer) such as dynamic DETR \cite{dai2021dynamic} and deformable DETR \cite{zhu2020deformable} utilize self-attention mechanisms to treat images as sequences of patches, which helps in integrating a global context and eliminates the need for Non-Maximum Suppression (NMS) \cite{hosang2017learning}, streamlining post-processing \cite{carion2020end}. Vision Language Models (VLMs) such as CLIP (Contrastive Language-Image Pre-training) represent an emerging area that integrates textual and visual data, aiming to enhance robustness through multimodal learning, though their application in real-world scenarios, particularly in robotics and automation, is still developing. On the other hand, Hybrid models such as RetinaMask, Sparse Coding models like Online Dictionary Learning, and Traditional Feature-based methods such as Histogram of Oriented Gradients (HOG) are increasingly considered obsolete \cite{ren2013histograms, xie2013discriminative}. These have been superseded by more advanced systems that provide not only greater accuracy but also the capability to perform in real-time, a critical requirement for latency-sensitive operations such as those found in modern agricultural settings. As object detection continues to evolve, the focus remains on technologies that combine high precision with efficient processing, positioning CNN and Transformer-based models as the current state-of-the-art in the field.

\begin{figure}[ht!]
\centering
\includegraphics[width=0.90\linewidth]{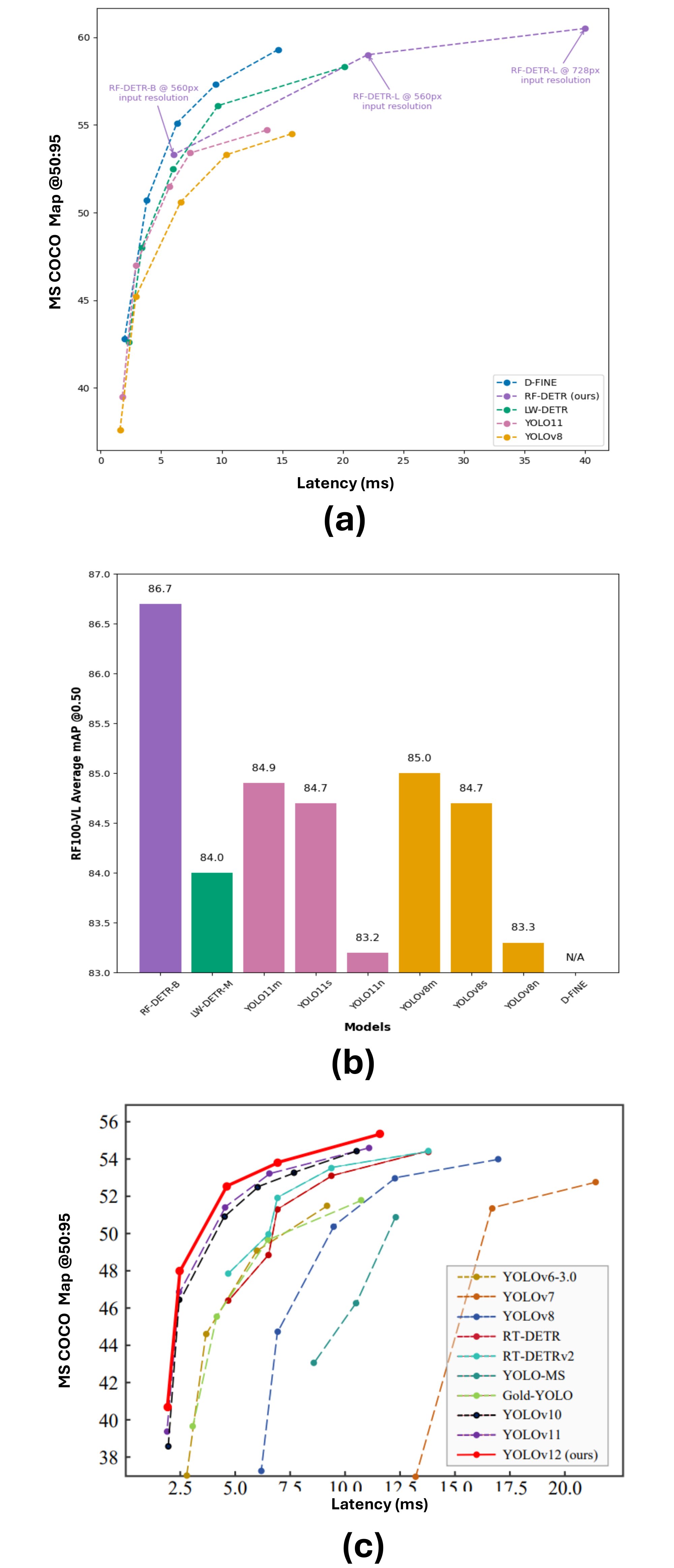}
\caption{\textbf{CNN vs Transformer-based model performance comparison focusing on YOLOv12 (CNN-based) and RF-DETR (Transformer-based) architectures: (a) RF-DETR object detection model benchmark evaluation with  YOLO11, YOLOv8 and other DETR-based object detection models ; (b)RF-DETR evaluation on the RF100-VL dataset, highlighting domain adaptability and edge deployment potential.   ; and (c) Performance overview of recent CNN-based models, includ- ing YOLOv6 through YOLOv12, Gold-YOLO, RT-DETR, RT-DETRv2, and YOLO-MS. b) RF-DETR benchmark results on the MS COCO dataset, surpassing 60\% mAP }}
\label{fig:background}
\end{figure}

Among the six primary object detection approaches illustrated in Figure \ref{fig:img1}, CNN-based and Transformer-based models have emerged as the most widely adopted and actively developed over the past five years. These two paradigms now dominate both research and real-world applications due to their scalability, accuracy, and adaptability. This ongoing dominance has sparked a competitive evolution between the two approaches, particularly with the release of powerful Transformer-based models such as RF-DETR, developed by Roboflow. RF-DETR integrates architectural innovations from Deformable DETR and LW-DETR, and utilizes a DINOv2 backbone, offering superior global context modeling and domain adaptability. The model eliminates reliance on anchor boxes and Non-Maximum Suppression (NMS), supporting end-to-end training and real-time inference. With two variants, Base (29M) and Large (128M), RF-DETR offers scalability from edge deployment to high-performance scenarios. The model has been shown to outperform YOLOv11 and is the only model to surpass 60\% mAP on the COCO dataset to date. Its performance on COCO and RF100-VL benchmarks is visualized in Figure \ref{fig:background}a and \ref{fig:background}b. However, despite its promise, RF-DETR has not yet been officially benchmarked against YOLOv12, the latest and most advanced model in the YOLO family. A comparative evaluation is warranted, particularly since YOLOv12 builds upon the strengths of YOLOv11, YOLOv10, and Gold-YOLO RT-DETR, as illustrated in Figure \ref{fig:background}c.

\subsection{CNN-based Object Detection Approaches} 
CNNs have been pivotal in advancing object detection, significantly since AlexNet catalyzed the field in 2012 \cite{o2015introduction}. These networks utilize hierarchical layers of convolutions, pooling, and nonlinear activations to effectively learn feature representations from images \cite{soydaner2022attention}. Unlike transformers that handle global relationships via attention mechanisms, CNNs are structured to excel at extracting local features, which is facilitated by their inherent inductive biases such as translation equivariance and the establishment of spatial hierarchies \cite{khan2023survey}. This fundamental architectural distinction makes CNNs particularly well-suited for scenarios demanding real-time processing and edge computing deployments, albeit with a noted limitation in their ability to model global contextual information comprehensively \cite{alzubaidi2021review, chen2018mobilefacenets}.

The progress in CNN architectures for object detection has been marked by several significant innovations as:
\begin{itemize}
    \item R-CNN Family: This series began with R-CNN in 2014 \cite{girshick2014rich}, which utilized selective search to generate region proposals that were then processed by CNNs to extract features, achieving a 53.3\% mAP on the PASCAL VOC dataset but with a high computational cost. Subsequent iterations, Fast R-CNN and Faster R-CNN, introduced ROI pooling and Region Proposal Networks (RPNs), respectively, enhancing the efficiency and speed of these models dramatically.

    \item Mask R-CNN: An extension of Faster R-CNN that includes a branch for predicting segmentation masks on each Region of Interest (ROI), effectively handling instance segmentation with a high precision level \cite{he2017mask, sapkota2024comparing}.

    \item YOLO Series: Starting with YOLOv1 \cite{redmon2016you}, which reframed object detection as a single regression problem from image pixels to bounding box coordinates and class probabilities, through to YOLOv12 \cite{tian2025yolov12}, which introduced improvements like anchor-free detection and dynamic label assignment for enhanced accuracy and efficiency \cite{sapkota2024yolov10, sapkota2025improved, meng2025yolov10, sapkota2024comparing}.

    \item SSD: This model combined multi-scale feature maps with default bounding boxes to perform detections, facilitating direct classification and localization from feature maps without needing separate region proposals \cite{liu2016ssd}.

    \item RetinaNet: Known for tackling class imbalance with the focal loss function, which helps focus the model on hard-to-classify examples by down-weighting the loss assigned to well-classified examples \cite{lin2017focal}.

    \item EfficientDet: This model utilized a scaling method that systematically adjusts the depth, width, and resolution of the network, integrated with a BiFPN for feature fusion across different scales, achieving high efficiency and accuracy \cite{tan2020efficientdet}.
\end{itemize}
\subsection{Transformers-based Object Detection Approaches} 
DETR have revolutionized object detection by integrating transformer architectures, traditionally used in natural language processing, into visual recognition tasks \cite{carion2020end}. Introduced by Facebook AI in 2020, DETR presents a novel approach by treating object detection as a direct set prediction problem, eliminating the need for traditional components like anchor boxes and complex post-processing steps such as Non-Maximum Suppression (NMS) \cite{carion2020end}. At its core, DETR uses a standard CNN backbone, typically ResNet-50, for initial feature extraction. This is followed by a transformer that consists of an encoder and a decoder where the encoder processes the spatial features across the image and the decoder uses learned object queries to predict the presence of objects along with their categories and bounding boxes, all in parallel.

The key architectural variants of DETR have addressed its initial shortcomings such as slow convergence and high computational demands:

\begin{itemize}
    \item Deformable DETR: Introduced to tackle the inefficiencies of the standard transformer attention mechanism, it employs deformable attention which focuses on a small set of key sampling points around each reference point, significantly reducing the computational load and improving detection of small objects \cite{zhu2020deformable}. This variant leverages iterative bounding box refinement and multi-scale features to enhance accuracy and speed up training .
    \item RT-DETR: Developed for real-time applications, this variant from Baidu features a hybrid encoder that merges CNN and transformer features to optimize both intra-scale interaction and cross-scale fusion, achieving impressive speeds on standard hardware. It introduces IoU-aware query selection, dynamically adjusting the decoding process based on predicted objectness scores \cite{wang2022farmland, lin2024dcea}.
    \item Co-DETR: Enhances training stability and performance by implementing a dual supervision strategy that combines traditional one-to-many (like Faster R-CNN) and one-to-one (like DETR) label matching \cite{zong2023detrs}. This approach, supported by hierarchical attention mechanisms, significantly improves feature representation, especially in challenging conditions such as occlusions \cite{zhang2025improved}.
    \item YOLOS: Stands out by adapting Vision Transformers (ViTs) directly for object detection without any CNNs \cite{fang2021you}. It uses a sequence of image patches (tokens) along with a set of learnable detection tokens, demonstrating that transformers can effectively encode spatial relationships inherent in detection tasks \cite{zhao2024detrs}.
    \item OWL-ViT: Expands the applicability of transformers to open-vocabulary detection by integrating vision and language, using a transformer decoder to align image features with text queries \cite{minderer2022simple}. This model facilitates zero-shot detection, where the system can recognize objects it has never seen during training, described only by text \cite{heigold2023video, wang2025effowt}.
    \item DINO (DETR with Improved Denoising Anchor boxes): Focuses on enhancing small object detection through a novel training strategy that involves adding noise to ground truth boxes and learning to predict corrective offsets, improving precision and robustness \cite{zhang2022dino}. 
    \item RF-DETR: Released by Roboflow, RF-DETR is a real-time transformer-based object detection model that achieves 60.5 mAP at 25 FPS on an NVIDIA T4 GPU, outperforming models like YOLOv11 and LW-DETR on benchmarks such as COCO and RF100-VL \cite{robicheauxroboflow100}. Its architecture is designed for high-speed edge deployment and domain adaptability, with two variants: RF-DETR-Base (29M parameters) and RF-DETR-Large (128M parameters). 
\end{itemize}

\subsection{Objectives}
Despite significant advances in object detection, the performance of state-of-the-art models in complex, label-ambiguous agricultural environments remains underexplored. The recently released RF-DETR, a Transformer-based real-time object detection model developed by Roboflow, has demonstrated remarkable performance by surpassing 60\% mAP on the MS COCO dataset, the highest recorded for any Transformer-based detector to date. However, RF-DETR’s performance has only been benchmarked against earlier versions of YOLO, including YOLOv11, and a few other models like LW-DETR, leaving a notable gap in comparative evaluation with YOLOv12, the most recent and advanced CNN-based detector from the YOLO family. This lack of direct comparison has created uncertainty regarding which model, RF-DETR or YOLOv12, offers better detection capabilities in real-world conditions, particularly under occlusion, camouflage, and ambiguous labeling.

This study addresses this gap by conducting a detailed evaluation of RF-DETR and YOLOv12 for the task of greenfruit detection in commercial apple orchards. Immature green apple fruitlets are critical for early yield estimation and thinning but are notoriously difficult to detect due to their small size, color similarity with the background, and frequent occlusion by foliage or other fruits. This visual complexity results in label ambiguity, making it difficult to determine whether fruitlets are fully visible, partially visible, or completely hidden conditions that challenge both manual annotation and automated detection.

To assess the robustness of these two architectures, we developed a custom dataset and evaluated both models using identical training protocols and hyperparameters. Performance was assessed in both single-class and multi-class detection tasks using key metrics: Precision, Recall, F1-Score, mAP@50, and mAP@50:95. We also measured inference speed and processing efficiency, aiming to provide a clear, evidence-driven comparison of CNN-based versus Transformer-based object detection in precision agriculture.

\begin{figure*}[t!]
\centering
\includegraphics[width=0.80\linewidth]{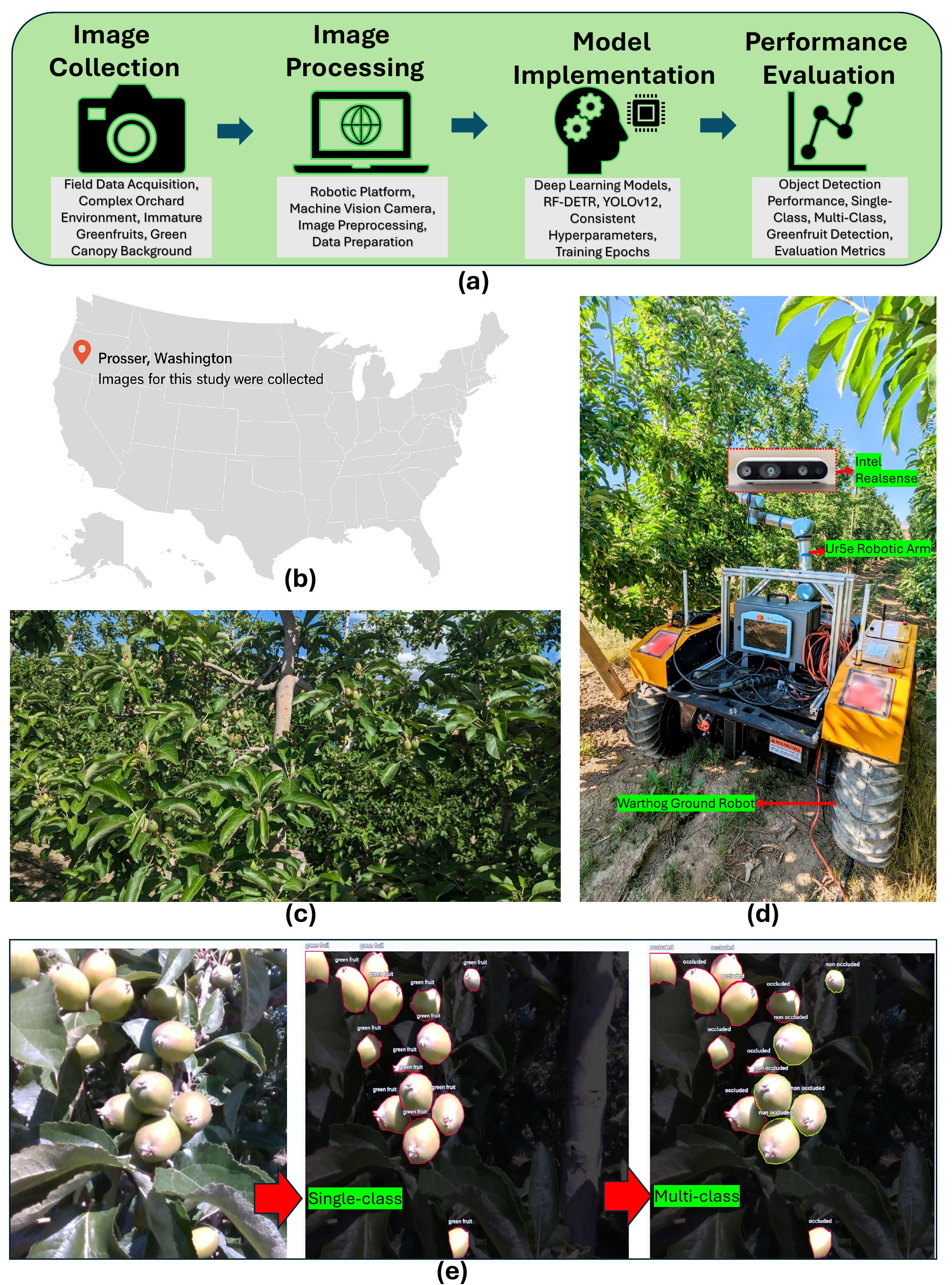}
\caption{\textbf{ Overview of data collection setup and environment: a) Flow diagram showing the methodology of RF-DETR vs YOLOv12 comparision ; b) Map highlighting the study location in Prosser, Washington, USA ; c) of 'Scifresh' apple trees, known as Jazz apples;  d) The robotic platform used for image acquisition, featuring an Intel RGB-D camera mounted on a UR5e robotic arm, capturing images of immature greenfruits in complex orchard environment.}}
\label{fig:method1}
\end{figure*}

\section{Methods}
This experiment is performed in four steps, as depicted in Figure \ref{fig:method1}a. Initially, real field images were collected from a commercial orchard under complex conditions, characterized by immature greenfruits camouflaged against a green canopy, presenting significant challenges for machine vision due to occlusions. Subsequently, these images were captured using a robotic platform and a machine vision camera, followed by preprocessing and preparation. In the third step, two deep learning models, RF-DETR and YOLOv12, were implemented using the same dataset, hyperparameters, and number of epochs. Finally, the performance of these models was evaluated in terms of their ability to detect single-class and multi-class greenfruit objects in this challenging orchard environment.

\subsection{Study Site and Data Acquisition}
Data acquisition for this study was conducted in a commercial orchard situated in Prosser, Washington State, USA, as illustrated in Figure \ref{fig:method1}b. The orchard was densely planted with 'Scifresh' apple trees, commonly known as Jazz apples. The selection of this specific orchard was due to its complex environmental conditions characterized by the green color of immature fruitlets blending with the green canopy background, as depicted in Figure \ref{fig:method1}c. This similarity in color created a challenging scenario for accurate image detection due to significant occlusions and visual confusion, typical in complex orchard scenes.

Image collection was executed using a sophisticated robotic platform that incorporated an Intel RGB-D camera, which was mounted on a UR5e robotic arm, as depicted in Figure \ref{fig:method1}d. This setup enabled the precise capture of RGB images, specifically focusing on the immature 'Scifresh' apple fruitlets. The imagery was collected in May 2024, just prior to the commencement of fruitlet thinning activities. The timing for the collection was carefully chosen based on continuous monitoring of the orchard's developmental stages (before thinning, exactly during the fruitlet thinning week) and in consultation with local growers and orchard workers to ensure optimal data relevance for the study.

The orchard, established in 2008, was methodically laid out with tree rows spaced 3 meters apart and an intra-row spacing of 1 meter. Throughout the course of this study, a total of 857 images were captured utilizing an Intel RealSense D435i camera, as shown in Figure \ref{fig:method1}d. The selected camera is equipped with a depth-sensing system that operates on active infrared (IR) stereo vision, complemented by an inertial measurement unit (IMU). This camera’s depth sensor employs structured light technology, which utilizes a pattern projector to induce disparities between stereo images captured by two IR cameras.

The camera's 3D sensor boasts a resolution of 1280 × 720 pixels, capable of capturing depth information up to a distance of 10 meters. It supports a frame rate of up to 90 frames per second (fps), and features a horizontal field of view (HFOV) of 69.4° and a vertical field of view (VFOV) of 42.5°. Additionally, the integrated 6-axis IMU provides critical orientation data, significantly enhancing the alignment of depth data with the actual scene, thus improving the overall understanding and analysis of the captured images. This detailed and methodical approach to data collection was fundamental in addressing the visual complexities presented by the orchard environment.

\subsection{Data Preprocessing and Preparation}
Following data collection, the acquired RGB images underwent a systematic preprocessing and annotation pipeline to prepare them for deep learning model training and evaluation. Image annotation was performed manually using the Roboflow platform (Roboflow, Des Moines, Iowa), a widely used tool for custom dataset generation in computer vision workflows. The dataset construction involved two labeling schemes: (i) a single-class dataset and (ii) a multi-class dataset, both of which aimed to capture the inherent complexity of greenfruit detection under real-world orchard conditions.

In the first scheme, all visible immature apples were annotated under a single class labeled as “greenfruit,” regardless of their degree of visibility or occlusion. A total of 857 high-resolution orchard images were uploaded and processed in Roboflow for this purpose. As illustrated in the center image of Figure \ref{fig:method1}e, this dataset captured a wide range of greenfruit appearances, resulting in 4,125 individual object labels. The uniform labeling in this scheme was suitable for establishing baseline detection performance but did not capture the dynamics of visual challenges like partial occlusions or background blending. To explore these complexities more explicitly, a second labeling scheme was developed to create a multi-class dataset. In this case, each greenfruit was categorized into one of two classes: occluded greenfruit and non-occluded greenfruit. The classification criteria were based on the degree of visibility. Greenfruits with at least 90\% of their surface area clearly visible, unobstructed by foliage, branches, or other fruits, were labeled as non-occluded. Conversely, any fruit partially hidden, whether by overlapping apples, intersecting leaves, or obstructing branches was labeled as occluded. This dynamic annotation approach is depicted in the rightmost image of Figure \ref{fig:method1}e.

However, the labeling process was complicated by label ambiguity, a critical issue in computer vision tasks, especially in natural environments. Label ambiguity refers to the uncertainty or subjectivity in assigning a label due to unclear visual boundaries, overlapping objects, or inconsistent visibility. In this study, several practical instances of label ambiguity emerged. First, in cases where multiple greenfruits clustered tightly, it was often unclear whether one was partially occluding the other or if they were side-by-side. Second, some fruits appeared occluded due to lighting and shadows rather than actual physical obstruction, leading to inconsistent labeling across images. Third, foliage sometimes mimicked the texture and color of immature fruits, making it difficult to distinguish between the object of interest and the background. Fourth, partial occlusions at the edges of images often left annotators uncertain whether to classify the object as occluded or simply truncated due to the field of view. These examples underscore why greenfruit detection in real orchard environments is particularly prone to labeling inconsistencies. Although classification guidelines were applied rigorously, the complex interplay between object geometry, environmental texture, and visibility made complete objectivity difficult to achieve. Thus, the term label ambiguity is used to describe the dataset's inherent subjectivity and the potential variability it introduces during model training and evaluation.
\begin{figure*}[ht!]
\centering
\includegraphics[width=0.83\linewidth]{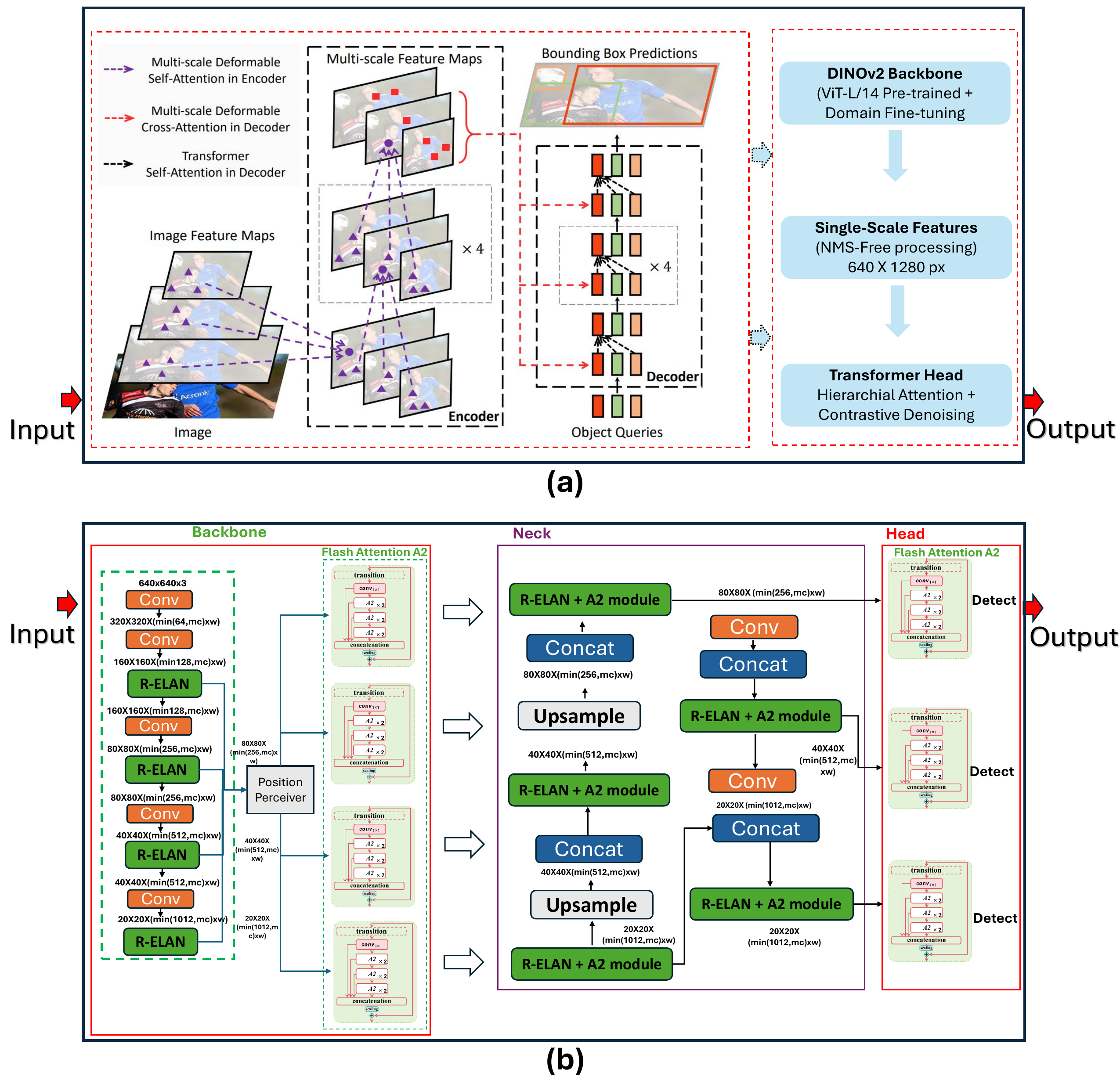}
\caption{\textbf{(a) RF-DETR Architecture diagram for object detection  ; (b) YOLOv12 Architecture Diagram for object detection}}
\label{fig:architectures}
\end{figure*}

\subsection{Training Object Detection Models}
\subsubsection{Training RF-DETR Object Detection Model}
RF-DETR is a real-time, transformer-based object detection architecture optimized for both accuracy and efficiency \footnote{\url{https://blog.roboflow.com/rf-detr/}} \footnote{\url{https://github.com/roboflow/rf-detr}}. As illustrated in Figure~\ref{fig:architectures}a, RF-DETR builds upon the foundations of Deformable DETR and LW-DETR, integrating a pre-trained DINOv2 vision transformer as its backbone. This backbone enhances cross-domain generalization through self-supervised learning, making the model highly adaptable to domain-specific challenges like greenfruit detection in agricultural environments.

A key innovation of RF-DETR is its ability to eliminate traditional object detection components such as anchor boxes and NMS. Instead, it uses a transformer-based encoder-decoder architecture with deformable cross-attention to selectively attend to spatially relevant features, improving detection under occlusion, clutter, and camouflage. Unlike traditional DETR variants, RF-DETR employs a single-scale feature extraction strategy to reduce computational overhead, thus enabling faster inference without compromising accuracy.

Two variants of the model are available: RF-DETR-Base (29 million parameters) and RF-DETR-Large (128 million parameters). In this study, the \textbf{**RF-DETR-Base**} model was selected due to its balance between computational efficiency and high detection accuracy, making it suitable for real-time processing in field robotics. The RF-DETR-Base model achieves a mAP of 53.3 on the COCO benchmark and 86.7 mAP@50 on the RF100-VL dataset, positioning it as one of the few models to exceed 60\% mAP@50:95 in real-time settings.

Training followed the official Roboflow implementation. The model was initialized with DINOv2-pretrained weights and trained using the AdamW optimizer with a learning rate of 1e-4 and batch size of 8 for 300 epochs. The training leveraged hybrid encoder optimizations inspired by RT-DETR and deformable attention mechanisms. Loss functions included cross-entropy for classification and a combination of L1 and GIoU losses for bounding box regression.

Additionally, contrastive denoising training was employed to improve detection robustness for partially visible and small objects. RF-DETR also adopted collaborative label assignments for stability in ambiguous annotation conditions, as well as multi-resolution input support (640–1280 px), allowing latency-accuracy trade-offs without retraining. This configuration made RF-DETR-Base a powerful and efficient model for detecting occluded and camouflaged immature greenfruits in complex orchard environments.

\subsubsection{Training YOLOv12 Object Detection Model}
YOLOv12 represents a transformative leap in CNN-based object detection, merging the efficiency of traditional convolutional architectures with attention-inspired mechanisms to address modern computer vision demands \cite{tian2025yolov12}. Departing from previous YOLO iterations, the model introduces R-ELAN (Residual Efficient Layer Aggregation Network) as its core backbone as depicted in Figure~\ref{fig:architectures}b, combining residual connections with multi-scale feature fusion to resolve gradient bottlenecks while enhancing feature reuse across network depths. A novel 7×7 separable convolution layer replaces standard 3×3 kernels, preserving spatial context with 60\% fewer parameters than conventional large-kernel convolutions while implicitly encoding positional relationships, effectively circumventing the need for explicit positional embeddings used in transformer-based detectors. The neck architecture integrates FlashAttention-optimized area attention, dividing feature maps into four horizontal/vertical regions for localized processing without sacrificing global context, achieving 40\% reduced memory overhead compared to standard self-attention implementations. These innovations enable state-of-the-art accuracy while maintaining real-time performance, with the YOLOv12-S variant outperforming RT-DETR-R18 in both speed (1.2× faster) and precision (62.1 vs 59.3 COCO mAP). The architecture further supports multi-task learning through unified prediction pathways, allowing simultaneous object detection, oriented bounding box (OBB) estimation, and instance segmentation via specialized heads—a first for the YOLO series. Hardware-aware optimizations ensure sub-10ms inference on edge devices, with the 12n variant (2.1M parameters) achieving 9.8ms latency while maintaining robust detection of sub-50px objects through lightweight MLP ratios (1.2-2.0 vs traditional 4.0) in task-specific heads.

YOLOv12’s architectural refinements optimize convolutional operations for contemporary hardware while introducing transformer-like capabilities through innovative attention hybrids. The area attention mechanism processes feature map segments independently through FlashAttention’s memory-efficient algorithms, enabling precise region-specific focus without the computational burden of full self-attention. This design philosophy extends to the model’s scalability, offering four configurations (12n/12s/12m/12x) ranging from 2.1M to 42M parameters to accommodate edge deployments (Jetson Nano) to cloud clusters (A100 GPUs). Unlike previous YOLO versions limited to axis-aligned detection, YOLOv12 introduces an OBB head with angle prediction capabilities, critical for aerial imagery and document analysis. Training stability is enhanced through block-level residual scaling in R-ELAN, which prevents feature degradation in deep networks while maintaining the single-pass efficiency that defines the YOLO series. Benchmark results demonstrate 4-8\% higher mAP than YOLOv11 across all variants, with the 12x model achieving 68.9 mAP on COCO, surpassing similarly sized transformer hybrids like DINO-DETR in small object detection tasks. The architecture’s separation of feature extraction (backbone) and attention-driven refinement (neck) allows targeted optimization, enabling the 12s variant to process 4K video streams at 45 FPS on an NVIDIA T4 GPU. By integrating the parameter efficiency of CNNs with the contextual awareness of attention mechanisms, YOLOv12 establishes a new standard for real-time vision systems, particularly in industrial applications requiring simultaneous detection, segmentation, and geometric prediction under strict latency constraints.

\subsection{Training Methodology} 
The training procedures for both deep learning models RF-DETR and YOLOv12 were carried out under identical experimental settings to ensure a fair and rigorous comparison. All training was conducted on a workstation equipped with an Intel(R) Core(TM) i9-10900K CPU @ 3.70GHz (10 cores, 20 threads), running Ubuntu 24.04.1, and supported by an NVIDIA RTX A5000 GPU with 24 GB VRAM. This high-performance hardware configuration ensured sufficient computational resources for training large-scale object detection models. The RF-DETR object detection model, specifically the Base variant, was trained for 50 epochs on the single-class greenfruit dataset and 100 epochs on the multi-class dataset. 

Notably, RF-DETR demonstrated rapid convergence in the single-class setting, with performance plateauing at under 20 epochs, highlighting the model’s efficient learning dynamics and its suitability for low-epoch training regimes. YOLOv12 models, including YOLOv12X, YOLOv12L, and YOLOv12N, were each trained for 100 epochs for both single-class and multi-class datasets to ensure convergence and optimal generalization. RF-DETR was implemented in PyTorch using Roboflow’s \texttt{rf-detr} framework, which integrates a Deformable DETR architecture with a pre-trained DINOv2 backbone to enhance global context modeling and cross-domain adaptability. The YOLOv12 models were trained using the official Ultralytics PyTorch framework, optimized for fast detection and efficient edge deployment. For both models, the input image resolution was standardized to 640×640 pixels, a resolution commonly adopted in orchard-based object detection tasks. 

The models training was performed using FP32 precision with a batch size of approximately 16 images per iteration. The software environment included CUDA 11.7+ and cuDNN 8.4+, ensuring full compatibility with GPU acceleration and deep learning libraries. This standardized setup enabled a reliable comparative evaluation of model accuracy, convergence behavior, and training efficiency across both transformer- and CNN-based architectures.

\subsection{Performance Evaluation}
To rigorously assess the capabilities of RF-DETR and YOLOv12 in identifying greenfruits in a complex orchard environment, a comprehensive evaluation was conducted using standardized metrics. Both models underwent training and testing under uniform conditions utilizing the same datasets, number of training epochs, learning rates, optimizers, and batch sizes to ensure an equitable comparison between the CNN-based YOLOv12 and Transformer-based RF-DETR architectures.

\vspace{0.2cm} \noindent \textbf{Detection Evaluation Metrics}

The evaluation metrics employed were Precision, Recall, F1-Score, mean Average Precision (mAP@50 and mAP@50:95), and mean Intersection over Union (mIoU). These metrics quantify the performance based on the interaction between predicted bounding boxes and ground truth annotations:

\begin{itemize} \item \textbf{True Positive (TP)}: A predicted bounding box correctly identifies a ground truth fruit with an Intersection over Union (IoU) $\geq$ the defined threshold (commonly 0.50). \item \textbf{False Positive (FP)}: A predicted bounding box either insufficiently overlaps with any ground truth box (IoU $<$ 0.50) or erroneously marks a non-existent object. \item \textbf{False Negative (FN)}: A real fruit is overlooked by the detection model, with no corresponding predicted box sufficiently overlapping it. \end{itemize}

The metrics were calculated as follows:

\begin{equation} \text{Precision} = \frac{TP}{TP + FP} \end{equation}

\begin{equation} \text{Recall} = \frac{TP}{TP + FN} \end{equation}

\begin{equation} \text{F1-Score} = 2 \cdot \frac{\text{Precision} \cdot \text{Recall}}{\text{Precision} + \text{Recall}} \end{equation}

Precision assesses the accuracy of the detected greenfruits, Recall gauges the completeness of the detection, and F1-Score balances both aspects to provide a single measure of model efficacy.

\vspace{0.2cm} \noindent \textbf{Intersection over Union (IoU) and Mean IoU (mIoU)}

\begin{equation} \text{IoU} = \frac{\text{Area of Overlap}}{\text{Area of Union}} = \frac{TP}{TP + FP + FN} \end{equation}

IoU quantifies the exactness of the overlap between the predicted and actual bounding boxes, crucial in scenarios with densely packed and overlapping fruits. mIoU averages the IoU across all detections to give a holistic measure of spatial accuracy.

\vspace{0.2cm} \noindent \textbf{mAP@50 and mAP@50:95}

\begin{equation} \text{mAP@50} = \frac{1}{N} \sum_{i=1}^{N} \text{AP}_i(\text{IoU} \geq 0.50) \end{equation}

\begin{equation} \text{mAP@50:95} = \frac{1}{10} \sum_{t=0.50}^{0.95} \text{mAP}_t \end{equation}

mAP@50 measures the mean Average Precision at an IoU threshold of 0.50, commonly used to evaluate detection effectiveness. mAP@50:95 averages the AP at ten IoU thresholds from 0.50 to 0.95 (in increments of 0.05), offering a rigorous assessment of a model's precision across a range of criteria from loose to strict overlaps, reflecting the model's robustness in both precise and approximate detection scenarios.

\vspace{0.2cm} \noindent \textbf{Application to Our Dataset}

For the single-class detection task, all greenfruits were uniformly considered, while the multi-class task also evaluated the accuracy of classifying fruits as occluded or non-occluded. Misclassifications of occlusion status were considered false positives, and undetected fruits, particularly those obscured by occlusion, were counted as false negatives.

\section{Results}
\begin{figure*}[t!]
\centering
\includegraphics[width=0.85\linewidth]{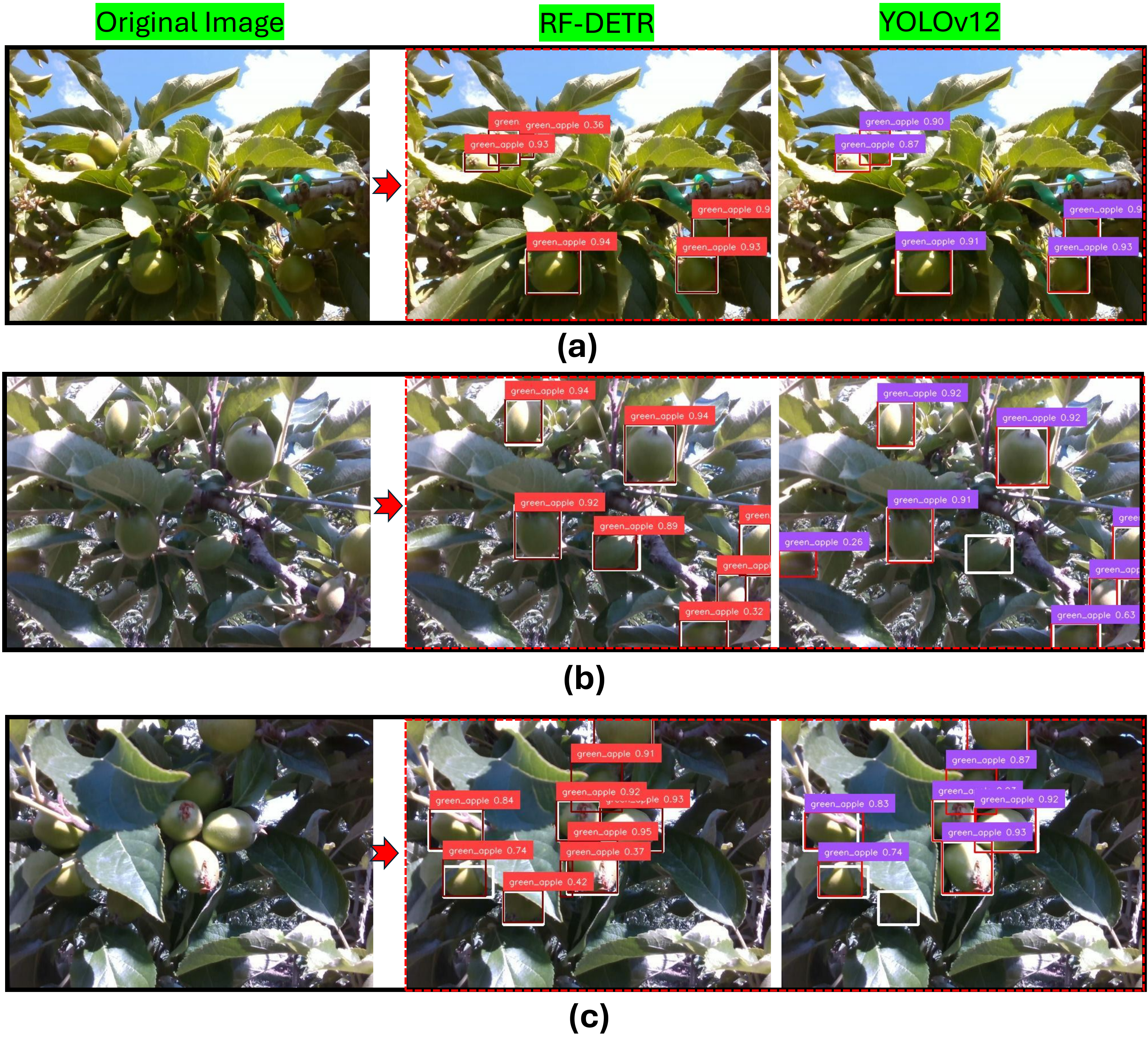}
\caption{\textbf{Visual comparison of single-class greenfruit detection using RF-DETR and YOLOv12 in complex orchard scenes. a) Three clustered greenfruits partially occluded by dense canopy; RF-DETR detected all, YOLOv12 missed one. b) A camouflaged greenfruit blending into the canopy; RF-DETR correctly detected it, YOLOv12 failed. c) A heavily occluded greenfruit with only the calyx visible under low light; RF-DETR identified it, YOLOv12 missed detection. }}
\label{fig:resultsingleclass}
\end{figure*}

\begin{figure*}[t!]
\centering
\includegraphics[width=0.85\linewidth]{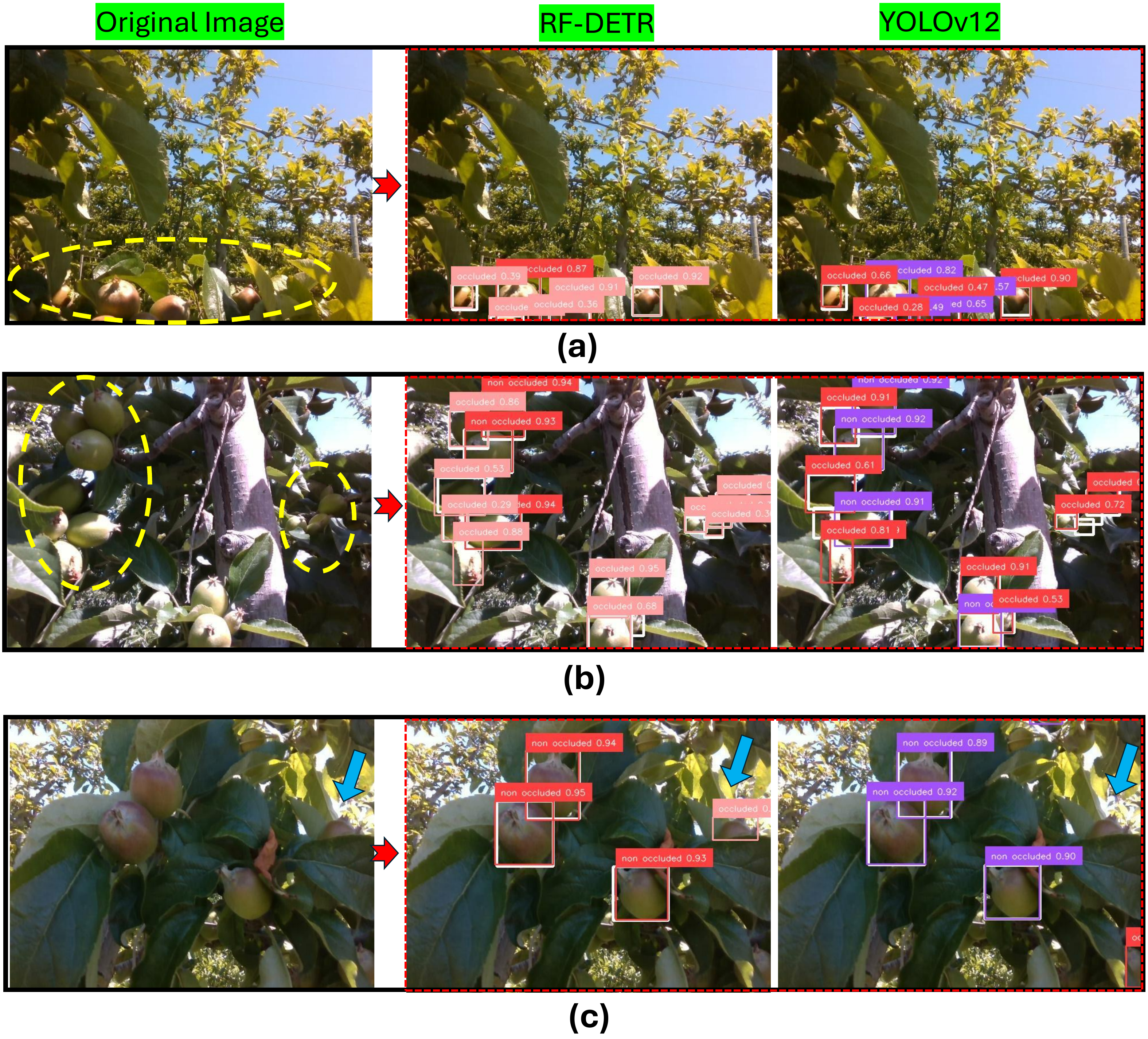}
\caption{\textbf{Visual comparison of multi-class greenfruit detection by RF-DETR and YOLOv12 under label ambiguity. (a) A dense fruit cluster at the image edge; YOLOv12 over-detected with false positives, while RF-DETR correctly detected 5 true greenfruits. (b) An occluded apple at the bottom; RF-DETR correctly labeled it as occluded, while YOLOv12 misclassified it as non-occluded. (c) A highly occluded fruit with only 10\% (approx) visibility; RF-DETR detected it as occluded, whereas YOLOv12 missed the detection entirely.}}
\label{fig:resultmulticlass}
\end{figure*}

The results for single-class and multi-class greenfruit detection using RF-DETR and YOLOv12 are presented to evaluate their performance in detecting green apples in complex orchard environments. Figure \ref{fig:resultsingleclass} displays three examples that highlight how each model performs in single-class detection scenarios, illustrating their efficacy in challenging conditions characterized by dense foliage and partial occlusions. Similarly, Figure \ref{fig:resultmulticlass} showcases three examples for multi-class detection, focusing on the models' ability to handle label ambiguity effectively. 

Each example includes the original RGB image collected in the orchard (left), detection output from RF-DETR (middle), and detection output from YOLOv12 (right). Key regions of interest are highlighted with yellow dotted circles, focusing on areas where fruitlets were either clustered, camouflaged, or heavily occluded. In Figure \ref{fig:resultsingleclass}a, three immature green apples appeared closely clustered within a dense canopy, with significant partial occlusions caused by overlapping leaves. The original image, shown on the left, presented a challenging scenario due to low fruit-background contrast and complex foliage structure. As illustrated in the middle image of Figure \ref{fig:resultsingleclass}a, RF-DETR successfully detected all three greenfruit instances, correctly bounding each fruit despite their partial visibility. In contrast, YOLOv12, shown on the right, detected only two out of the three apples and failed to identify the third fruit, which was most heavily occluded. This result highlighted RF-DETR's superior capability in handling complex spatial relationships and occlusions. Figure \ref{fig:resultsingleclass}b provided another challenging condition where a single green apple within the yellow dotted circle was camouflaged due to its visual similarity with the surrounding canopy. Despite the low contrast between the fruit and the background, RF-DETR accurately identified the greenfruit, as shown in the middle figure. Conversely, YOLOv12 failed to detect this fruit, indicating its limitations in distinguishing camouflaged targets in homogeneous backgrounds.

In Figure \ref{fig:resultsingleclass}c, a different scenario was examined where only a small portion (approximately 10\%) of the fruit’s calyx was visible due to heavy occlusion by a leaf and low ambient lighting. The original RGB image (left) demonstrated minimal visible surface area of the fruit. Remarkably, RF-DETR still managed to detect the partially exposed fruit in the middle image, whereas YOLOv12 again failed to register the object in its detection output. These examples consistently demonstrated RF-DETR’s higher sensitivity and robustness in single-class greenfruit detection, especially under conditions of occlusion, camouflage, and low visibility, compared to the CNN-based YOLOv12 model.

Figure \ref{fig:resultmulticlass} presents qualitative comparisons between RF-DETR and YOLOv12 for multi-class greenfruit detection, where fruitlets were classified as either occluded or non-occluded. This evaluation highlights the models’ performance in handling label ambiguity—situations where visibility is unclear due to clustering, occlusion, or edge truncation. 

Likewise, in Figure \ref{fig:resultmulticlass}a, a dense cluster of greenfruits appears near the image edge, creating a highly ambiguous scene. As shown in the rightmost image, YOLOv12 detected 7 greenfruit instances in this region. However, ground truth annotation confirmed that only 5 greenfruits were actually present. YOLOv12 misclassified background textures or overlapping canopy features as non-occluded apples, resulting in false positives. In contrast, RF-DETR, shown in the middle image, correctly detected the 5 actual greenfruits but missed classifying them into occluded/non-occluded categories with high certainty. In this example, YOLOv12 appeared visually more active but less accurate, while RF-DETR provided precise detection with lower misclassifications.

Furthermore, in Figure \ref{fig:resultmulticlass} b, yellow circles in the original image (left) highlight true greenfruits. RF-DETR detected 12 apples, including an occluded apple at the bottom of the frame, which was correctly labeled occluded (middle). YOLOv12 detected 11 apples but incorrectly labeled the bottom occluded apple as non-occluded (right), suggesting that RF-DETR object detection model performed better in differentiating occlusion classes, likely due to its global attention modeling. 

Likewise, Figure \ref{fig:resultmulticlass}c presents a challenging low-visibility case, where only 10\% of a greenfruit is visible beneath leaf cover (indicated by a blue arrow). RF-DETR successfully detected and classified it as occluded (middle), while YOLOv12 failed to detect the fruit at all (right). This reinforces RF-DETR’s strength in handling extreme occlusion.

\subsection{Evaluation of Precision, Recall and F1-Score}
Among all the models evaluated, YOLOv12N achieved the highest performance in terms of recall (0.8901) and the F1 score (0.8784) for single-class greenfruit detection, indicating its strong ability to detect almost all greenfruit instances while maintaining balanced precision. However, in terms of precision, YOLOv12L outperformed all other configurations of YOLOv12 and RF-DETR object detection model, achieving a top value of 0.8892 in single-class detection. This demonstrates the superior ability of YOLOv12L in reducing false positives and making accurate predictions. Detailed precision, recall, and F1 metrics for all models and detection types are presented in Table \ref{tab:model_perf_comparison}.

\begin{table*}[!ht]
\caption{\textbf{Comparative analysis of Precision, Recall, and F1 Score for single-class and multi-class greenfruit detection using RF-DETR (Transformer-based) and YOLOv12 (CNN-based) object detection algorithms. The table presents model performance across different YOLOv12 configurations (X, L, N) and highlights their effectiveness in detecting greenfruits under varying complexity and class conditions in orchard environments.}}
\label{tab:model_perf_comparison}
\centering
\begin{tabular}{lcccccc}
\toprule
\multirow{2}{*}{\textbf{Models}} & \multicolumn{3}{c}{\textbf{Single-Class}} & \multicolumn{3}{c}{\textbf{Multi-Class}} \\
\cmidrule(lr){2-4} \cmidrule(lr){5-7}
 & \textbf{Precision} & \textbf{Recall} & \textbf{F1 Score} & \textbf{Precision} & \textbf{Recall} & \textbf{F1 Score} \\
\midrule
RF-DETR     & 0.8663 & 0.8828 & 0.8744 & 0.7652 & 0.8109 & 0.7874 \\
YOLOv12X    & 0.8797 & 0.8595 & 0.8694 & 0.6986 & 0.8261 & 0.7570 \\
YOLOv12L    & 0.8892 & 0.8631 & 0.8759 & 0.7692 & 0.7827 & 0.7759 \\
YOLOv12N    & 0.8671 & 0.8901 & 0.8784 & 0.7569 & 0.7406 & 0.7487 \\
\bottomrule
\end{tabular}
\end{table*}

\begin{figure*}[t!]
\centering
\includegraphics[width=0.78\linewidth]{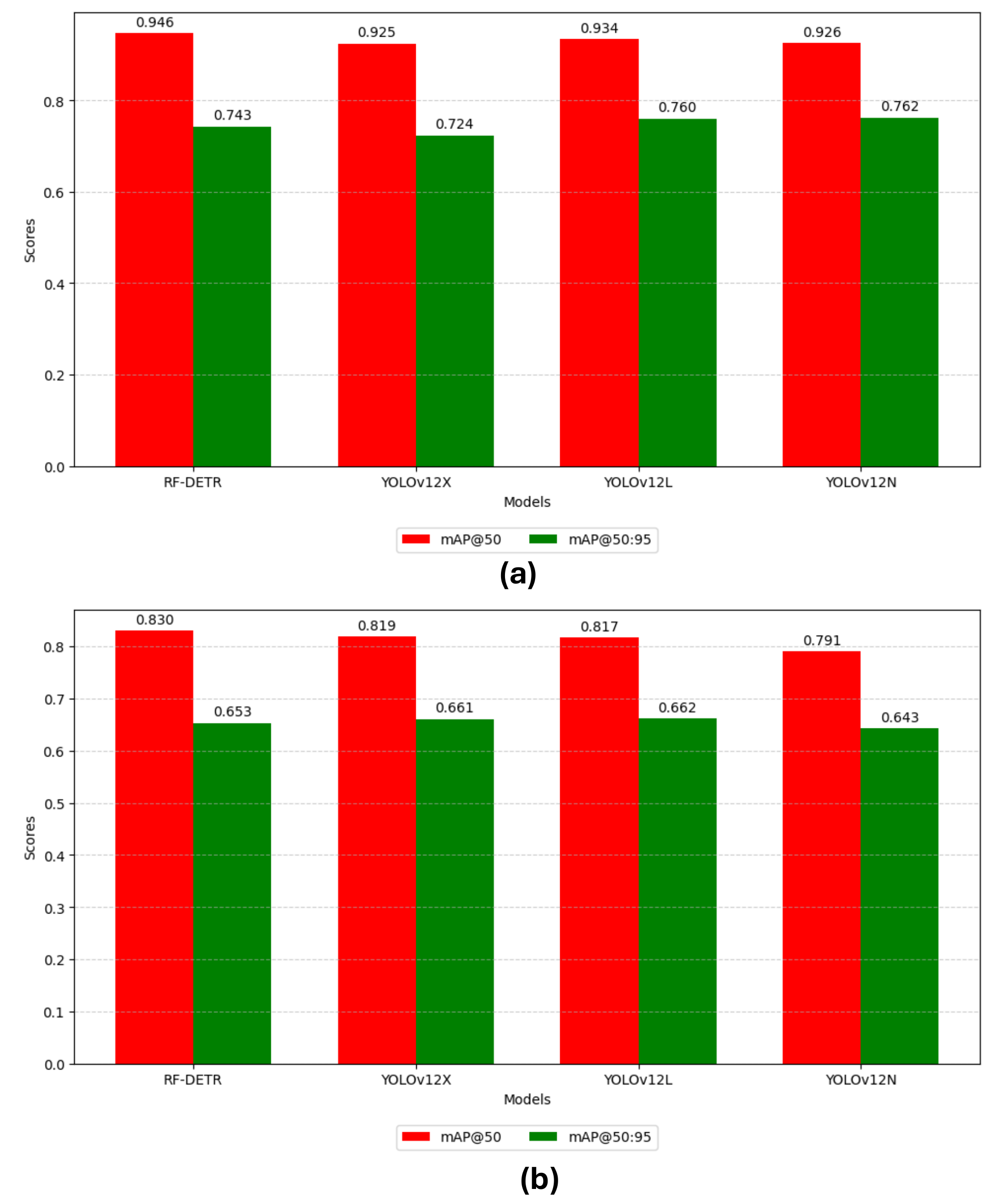}
\caption{\textbf{Mean Average Precision (mAP) comparison for greenfruit detection using  RF-DETR and YOLOv12 object detection models: a) mAP@50 for single-class detection. b) mAP@50 and mAP@50:95 for multi-class detection }}
\label{fig:maps}
\end{figure*}
\begin{figure*}[t!]
\centering
\includegraphics[width=0.75\linewidth]{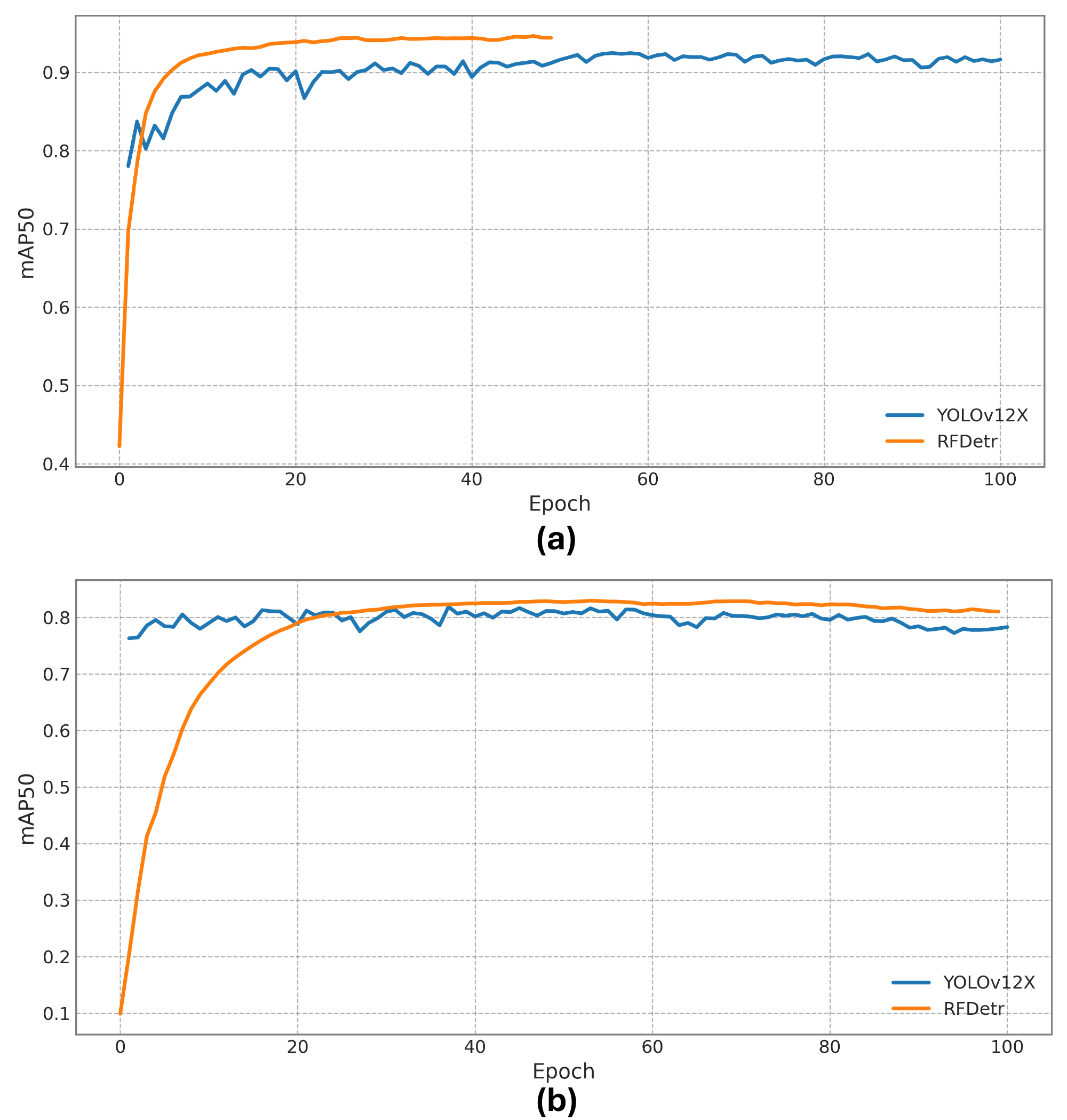}
\caption{\textbf{Training Dynamics and Model Convergence Analysis: mAP@50 vs. Epoch Curves for Object Detection Models. (a) Single-class greenfruit detection showing the performance trajectory of RF-DETR and YOLOv12X models over training epochs. (b) Multi-class greenfruit detection comparing the convergence patterns of both models throughout the training period}}
\label{fig:mapepoch}
\end{figure*}
\subsection{Analysis of Mean Average Precision (mAP)}
For single-class greenfruit detection, RF-DETR outperformed all other models with the highest mAP@50 of 0.9464, indicating its superior ability to accurately detect and localize greenfruits with sufficient overlap. Furthermore, RF-DETR achieved a mAP@50:95 of 0.7433, the second-highest among the tested models. Although YOLOv12N achieved a slightly higher mAP@50:95 of 0.7620, RF-DETR’s consistently higher mAP@50 suggests more reliable performance in practical orchard detection scenarios, especially when bounding box precision at the 50\% threshold is critical. In the multi-class detection scenario, where greenfruits were labeled as either occluded or non-occluded, YOLOv12L achieved the highest mAP@50:95 of 0.6622, slightly outperforming YOLOv12X and RF-DETR object detection, which achieved 0.6609 and 0.6530 respectively. This suggests that YOLOv12L was marginally better in maintaining detection consistency across varying degrees of overlap under label ambiguity. However, RF-DETR object detection model achieved the highest mAP@50 of 0.8298 in the multi-class setting, confirming its strength in confidently detecting objects with at least 50\% spatial alignment. These findings indicate that RF-DETR excels in spatially accurate detections, particularly for clearly visible fruits, while YOLOv12L performs slightly better under complex classification scenarios involving occlusion. The complete visualization of these metrics is presented in Figure \ref{fig:maps}, where Figure \ref{fig:maps}a shows the mAP@50 for single-class detection, and Figure \ref{fig:maps}b illustrates mAP@50 and mAP@50:95 for multi-class detection.

\subsection{Training Dynamics and Model Convergence Analysis}
Figure \ref{fig:mapepoch} provides a detailed visualization of the mean Average Precision (mAP@50) against the number of training epochs for both RF-DETR and YOLOv12X models, shedding light on their learning efficiency and stability during the training phase. In Figure \ref{fig:mapepoch}a, which tracks performance for single-class greenfruit detection, RF-DETR, a transformer-based object detection model, demonstrates an impressive early convergence, plateauing before 10 epochs. This rapid stabilization underscores RF-DETR's swift adaptability to complex orchard scenes, a significant advance over traditional CNN-based models like YOLOv12X.

Similarly, Figure \ref{fig:mapepoch}b illustrates the training progression for multi-class detection scenarios. Here, RF-DETR also shows superior convergence, reaching stability at around 20 epochs, far sooner than its CNN counterpart, which continues to seek equilibrium. This faster convergence of RF-DETR in both single and multi-class settings is emblematic of the inherent strengths of transformer technology in handling dynamic and visually cluttered environments efficiently.

These following five observations highlight several key advantages of employing transformer-based models like RF-DETR in object detection tasks:
\begin{enumerate}
    \item \textbf{Accelerated Learning Curve:} RF-DETR’s ability to reach peak performance quickly reduces the computational resources and time required for training, enhancing productivity and reducing operational costs.
    \item \textbf{Stable Performance:} The model maintains consistent accuracy over time, indicating robustness against overfitting and the ability to generalize well from limited epoch training.
    \item \textbf{Adaptability:} RF-DETR's architecture is evidently well-suited for complex detection environments, such as those in precision agriculture, where occlusions and varying object appearances prevail.
    \item \textbf{Efficient Resource Utilization:} By converging quickly, RF-DETR maximizes the utility of computational resources, allowing for more tasks to be performed in the same computational budget.
    \item \textbf{Edge Deployment:} The model's quick adaptation and stable performance make it ideal for deployment in edge devices where computational resources and power are limited.
\end{enumerate}

\section{Discussion}
The progression of greenfruit detection technologies is closely aligned with recent advancements in computer vision, where each new model iteration introduces more nuanced capabilities, especially in complex agricultural settings. Notable contributions include those by \cite{sapkota2024comparing, sapkota2024zero}, which provided a comparative analysis of YOLOv11 and YOLOv8, focusing on their efficacy in segmenting occluded and non-occluded immature green fruits. Similarly, \cite{sapkota2024immature} explored size estimation techniques using YOLOv8 combined with geometric shape fitting on 3D point cloud data, aiming to enhance yield predictions and crop management decisions. These studies underscore the ongoing efforts to refine the accuracy and efficiency of detection systems in variable orchard environments.

In this context, our study leverages the RF-DETR model, which has set new benchmarks in detection performance. RF-DETR's transformer-based architecture has achieved a mAP@50 of 0.9464, surpassing YOLOv12 and demonstrating superior spatial detection accuracy, particularly under conditions of partial visibility and camouflage \cite{sapkota2024yolo11}. This model's efficiency is further highlighted by its rapid convergence during training, indicating a significant advancement over traditional CNN-based models.

The integration of Vision-Language Models (VLMs) and open-vocabulary detection also represents a pivotal shift towards more dynamic and adaptable detection systems. These technologies, as reviewed in \cite{liu2024green, sapkota2025multimodal}, allow for the identification of a broader range of fruit types and characteristics without retraining. This adaptability is crucial for managing the diverse conditions typical of agricultural settings, where fruit appearances and environmental factors such as lighting and occlusion vary considerably.
the application of multimodal learning approaches that incorporate various sensory data types promises to resolve longstanding challenges such as camouflage and label ambiguity. The exploration of semi-supervised and few-shot learning paradigms could reduce reliance on extensive labeled datasets, facilitating quicker adaptation to new orchard environments \cite{liu2024mae}. Furthermore, the deployment of lightweight transformer variants and efficient VLMs for real-time field applications will be instrumental. These advancements will enable the development of mobile or edge-based systems that offer real-time analytics, crucial for immediate agricultural decision-making \cite{lv2024fcae}.  Continued advancements in these areas will undoubtedly forge detection systems that are not only highly accurate but also capable of semantic and contextual understanding. Such systems will drive the next wave of innovations in precision agriculture, ensuring that detection technologies are not only effective but also robust and adaptable to the complex dynamics of natural orchard environments.

\section{Conclusion} 
This study provided an in-depth evaluation of RF-DETR (Transformer-based) and YOLOv12 (CNN-based) object detection models for detecting greenfruits in commercial orchards with complex visual environments. The research process included gathering real-world images, preparing datasets with occlusion-based labels for both single-class and multi-class detection, and assessing the models under standardized conditions. The comparison was based on precision, recall, F1-score, and mean average precision (mAP@50 and mAP@50:95). The analysis extended to training dynamics, revealing that RF-DETR demonstrated a quicker convergence, achieving stable performance in fewer epochs compared to YOLOv12. This insight highlights RF-DETR's effectiveness in adapting to the variable conditions of orchard environments while maintaining accuracy across extensive training phases.

Key Findings:
\begin{itemize} 
\item \textbf{Single-Class Detection}: RF-DETR object detection model showcased superior performance with the highest mAP@50 of 0.9464, effectively localizing and detecting greenfruits amidst complex backgrounds. While YOLOv12N achieved the highest mAP@50:95 of 0.7620, RF-DETR remained consistently more accurate in cluttered and occluded scenarios. 
\item \textbf{Multi-Class Detection}: RF-DETR excelled in distinguishing between occluded and non-occluded fruits, registering the highest mAP@50 of 0.8298. YOLOv12L performed marginally better in mAP@50:95 with 0.6622, showcasing enhanced classification accuracy in detailed occlusion conditions.
\item \textbf{Model Training Dynamics and Convergence}: The RF-DETR object detection model was notable for its rapid training convergence, particularly in single-class scenarios where it plateaued at under 10 epochs, demonstrating both the efficiency and robustness of transformer-based architectures in handling dynamic visual data. 
\end{itemize}

\section*{Acknowledgment}This research is funded by the National Science Foundation and the United States Department of Agriculture, National Institute of Food and Agriculture through the “AI Institute for Agriculture” Program (Award No.AWD003473). We extend our heartfelt gratitude to Zhichao Meng, Astrid Wimmer, Randall Cason, Diego Lopez, Giulio Diracca for the data preparation; Martin Churuvija and Priyanka Upadhyaya for their invaluable efforts in logistical support during the data collection throughout this project. Special thanks to Dave Allan for granting commercial orchard access for this research experiment.
\section*{Declarations}
The authors declare no conflicts of interest.
\bibliographystyle{elsarticle-num-names}  
\bibliography{example}

\begin{thebibliography}{55}
\expandafter\ifx\csname natexlab\endcsname\relax\def\natexlab#1{#1}\fi
\providecommand{\url}[1]{\texttt{#1}}
\providecommand{\href}[2]{#2}
\providecommand{\path}[1]{#1}
\providecommand{\DOIprefix}{doi:}
\providecommand{\ArXivprefix}{arXiv:}
\providecommand{\URLprefix}{URL: }
\providecommand{\Pubmedprefix}{pmid:}
\providecommand{\doi}[1]{\href{http://dx.doi.org/#1}{\path{#1}}}
\providecommand{\Pubmed}[1]{\href{pmid:#1}{\path{#1}}}
\providecommand{\bibinfo}[2]{#2}
\ifx\xfnm\relax \def\xfnm[#1]{\unskip,\space#1}\fi
\bibitem[{Masmoudi et~al.(2019)Masmoudi, Ghazzai, Frikha, and Massoud}]{masmoudi2019object}
\bibinfo{author}{M.~Masmoudi}, \bibinfo{author}{H.~Ghazzai}, \bibinfo{author}{M.~Frikha}, \bibinfo{author}{Y.~Massoud},
\newblock \bibinfo{title}{Object detection learning techniques for autonomous vehicle applications},
\newblock in: \bibinfo{booktitle}{2019 IEEE international conference on vehicular electronics and safety (ICVES)}, \bibinfo{organization}{IEEE}, \bibinfo{year}{2019}, pp. \bibinfo{pages}{1--5}.
\bibitem[{Hnewa and Radha(2020)}]{hnewa2020object}
\bibinfo{author}{M.~Hnewa}, \bibinfo{author}{H.~Radha},
\newblock \bibinfo{title}{Object detection under rainy conditions for autonomous vehicles: A review of state-of-the-art and emerging techniques},
\newblock \bibinfo{journal}{IEEE Signal Processing Magazine} \bibinfo{volume}{38} (\bibinfo{year}{2020}) \bibinfo{pages}{53--67}.
\bibitem[{Elakkiya et~al.(2021)Elakkiya, Subramaniyaswamy, Vijayakumar, and Mahanti}]{elakkiya2021cervical}
\bibinfo{author}{R.~Elakkiya}, \bibinfo{author}{V.~Subramaniyaswamy}, \bibinfo{author}{V.~Vijayakumar}, \bibinfo{author}{A.~Mahanti},
\newblock \bibinfo{title}{Cervical cancer diagnostics healthcare system using hybrid object detection adversarial networks},
\newblock \bibinfo{journal}{IEEE Journal of Biomedical and Health Informatics} \bibinfo{volume}{26} (\bibinfo{year}{2021}) \bibinfo{pages}{1464--1471}.
\bibitem[{Mishra and Saroha(2016)}]{mishra2016study}
\bibinfo{author}{P.~K. Mishra}, \bibinfo{author}{G.~Saroha},
\newblock \bibinfo{title}{A study on video surveillance system for object detection and tracking},
\newblock in: \bibinfo{booktitle}{2016 3rd international conference on computing for sustainable global development (INDIACom)}, \bibinfo{organization}{IEEE}, \bibinfo{year}{2016}, pp. \bibinfo{pages}{221--226}.
\bibitem[{Badgujar et~al.(2024)Badgujar, Poulose, and Gan}]{badgujar2024agricultural}
\bibinfo{author}{C.~M. Badgujar}, \bibinfo{author}{A.~Poulose}, \bibinfo{author}{H.~Gan},
\newblock \bibinfo{title}{Agricultural object detection with you only look once (yolo) algorithm: A bibliometric and systematic literature review},
\newblock \bibinfo{journal}{Computers and Electronics in Agriculture} \bibinfo{volume}{223} (\bibinfo{year}{2024}) \bibinfo{pages}{109090}.
\bibitem[{Sa et~al.(2016)Sa, Ge, Dayoub, Upcroft, Perez, and McCool}]{sa2016deepfruits}
\bibinfo{author}{I.~Sa}, \bibinfo{author}{Z.~Ge}, \bibinfo{author}{F.~Dayoub}, \bibinfo{author}{B.~Upcroft}, \bibinfo{author}{T.~Perez}, \bibinfo{author}{C.~McCool},
\newblock \bibinfo{title}{Deepfruits: A fruit detection system using deep neural networks},
\newblock \bibinfo{journal}{sensors} \bibinfo{volume}{16} (\bibinfo{year}{2016}) \bibinfo{pages}{1222}.
\bibitem[{Singh and Krishnamurthi(2024)}]{singh2024iot}
\bibinfo{author}{P.~Singh}, \bibinfo{author}{R.~Krishnamurthi},
\newblock \bibinfo{title}{Iot-based real-time object detection system for crop protection and agriculture field security},
\newblock \bibinfo{journal}{Journal of Real-Time Image Processing} \bibinfo{volume}{21} (\bibinfo{year}{2024}) \bibinfo{pages}{106}.
\bibitem[{Yang et~al.(2024)Yang, Noguchi, and Hoshino}]{yang2024development}
\bibinfo{author}{L.~Yang}, \bibinfo{author}{T.~Noguchi}, \bibinfo{author}{Y.~Hoshino},
\newblock \bibinfo{title}{Development of a pumpkin fruits pick-and-place robot using an rgb-d camera and a yolo based object detection ai model},
\newblock \bibinfo{journal}{Computers and Electronics in Agriculture} \bibinfo{volume}{227} (\bibinfo{year}{2024}) \bibinfo{pages}{109625}.
\bibitem[{Gu et~al.(2018)Gu, Wang, Kuen, Ma, Shahroudy, Shuai, Liu, Wang, Wang, Cai et~al.}]{gu2018recent}
\bibinfo{author}{J.~Gu}, \bibinfo{author}{Z.~Wang}, \bibinfo{author}{J.~Kuen}, \bibinfo{author}{L.~Ma}, \bibinfo{author}{A.~Shahroudy}, \bibinfo{author}{B.~Shuai}, \bibinfo{author}{T.~Liu}, \bibinfo{author}{X.~Wang}, \bibinfo{author}{G.~Wang}, \bibinfo{author}{J.~Cai}, et~al.,
\newblock \bibinfo{title}{Recent advances in convolutional neural networks},
\newblock \bibinfo{journal}{Pattern recognition} \bibinfo{volume}{77} (\bibinfo{year}{2018}) \bibinfo{pages}{354--377}.
\bibitem[{Nimma and Zhou(2024)}]{nimma2024intelpvt}
\bibinfo{author}{D.~Nimma}, \bibinfo{author}{Z.~Zhou},
\newblock \bibinfo{title}{Intelpvt: intelligent patch-based pyramid vision transformers for object detection and classification},
\newblock \bibinfo{journal}{International Journal of Machine Learning and Cybernetics} \bibinfo{volume}{15} (\bibinfo{year}{2024}) \bibinfo{pages}{1767--1778}.
\bibitem[{Liu et~al.(2025)Liu, Zhan, Sun, Mao, and Wu}]{liu2025transformer}
\bibinfo{author}{H.~Liu}, \bibinfo{author}{Y.~Zhan}, \bibinfo{author}{J.~Sun}, \bibinfo{author}{Q.~Mao}, \bibinfo{author}{T.~Wu},
\newblock \bibinfo{title}{A transformer-based model with feature compensation and local information enhancement for end-to-end pest detection},
\newblock \bibinfo{journal}{Computers and Electronics in Agriculture} \bibinfo{volume}{231} (\bibinfo{year}{2025}) \bibinfo{pages}{109920}.
\bibitem[{Zang et~al.(2025)Zang, Li, Han, Zhou, and Loy}]{zang2025contextual}
\bibinfo{author}{Y.~Zang}, \bibinfo{author}{W.~Li}, \bibinfo{author}{J.~Han}, \bibinfo{author}{K.~Zhou}, \bibinfo{author}{C.~C. Loy},
\newblock \bibinfo{title}{Contextual object detection with multimodal large language models},
\newblock \bibinfo{journal}{International Journal of Computer Vision} \bibinfo{volume}{133} (\bibinfo{year}{2025}) \bibinfo{pages}{825--843}.
\bibitem[{Fu et~al.(2025)Fu, Yang, Mo, Yan, Wei, Meng, Xie, and Zheng}]{fu2025llmdet}
\bibinfo{author}{S.~Fu}, \bibinfo{author}{Q.~Yang}, \bibinfo{author}{Q.~Mo}, \bibinfo{author}{J.~Yan}, \bibinfo{author}{X.~Wei}, \bibinfo{author}{J.~Meng}, \bibinfo{author}{X.~Xie}, \bibinfo{author}{W.-S. Zheng},
\newblock \bibinfo{title}{Llmdet: Learning strong open-vocabulary object detectors under the supervision of large language models},
\newblock \bibinfo{journal}{arXiv preprint arXiv:2501.18954}  (\bibinfo{year}{2025}).
\bibitem[{Fu et~al.(2019)Fu, Shvets, and Berg}]{fu2019retinamask}
\bibinfo{author}{C.-Y. Fu}, \bibinfo{author}{M.~Shvets}, \bibinfo{author}{A.~C. Berg},
\newblock \bibinfo{title}{Retinamask: Learning to predict masks improves state-of-the-art single-shot detection for free},
\newblock \bibinfo{journal}{arXiv preprint arXiv:1901.03353}  (\bibinfo{year}{2019}).
\bibitem[{Tan et~al.(2020)Tan, Pang, and Le}]{tan2020efficientdet}
\bibinfo{author}{M.~Tan}, \bibinfo{author}{R.~Pang}, \bibinfo{author}{Q.~V. Le},
\newblock \bibinfo{title}{Efficientdet: Scalable and efficient object detection},
\newblock in: \bibinfo{booktitle}{Proceedings of the IEEE/CVF conference on computer vision and pattern recognition}, \bibinfo{year}{2020}, pp. \bibinfo{pages}{10781--10790}.
\bibitem[{Redmon et~al.(2016)Redmon, Divvala, Girshick, and Farhadi}]{redmon2016you}
\bibinfo{author}{J.~Redmon}, \bibinfo{author}{S.~Divvala}, \bibinfo{author}{R.~Girshick}, \bibinfo{author}{A.~Farhadi},
\newblock \bibinfo{title}{You only look once: Unified, real-time object detection},
\newblock in: \bibinfo{booktitle}{Proceedings of the IEEE conference on computer vision and pattern recognition}, \bibinfo{year}{2016}, pp. \bibinfo{pages}{779--788}.
\bibitem[{Sapkota et~al.(2024)Sapkota, Meng, Churuvija, Du, Ma, and Karkee}]{sapkota2024comprehensive}
\bibinfo{author}{R.~Sapkota}, \bibinfo{author}{Z.~Meng}, \bibinfo{author}{M.~Churuvija}, \bibinfo{author}{X.~Du}, \bibinfo{author}{Z.~Ma}, \bibinfo{author}{M.~Karkee},
\newblock \bibinfo{title}{Comprehensive performance evaluation of yolo11, yolov10, yolov9 and yolov8 on detecting and counting fruitlet in complex orchard environments},
\newblock \bibinfo{journal}{arXiv preprint arXiv:2407.12040}  (\bibinfo{year}{2024}).
\bibitem[{He et~al.(2017)He, Gkioxari, Doll{\'a}r, and Girshick}]{he2017mask}
\bibinfo{author}{K.~He}, \bibinfo{author}{G.~Gkioxari}, \bibinfo{author}{P.~Doll{\'a}r}, \bibinfo{author}{R.~Girshick},
\newblock \bibinfo{title}{Mask r-cnn},
\newblock in: \bibinfo{booktitle}{Proceedings of the IEEE international conference on computer vision}, \bibinfo{year}{2017}, pp. \bibinfo{pages}{2961--2969}.
\bibitem[{Dai et~al.(2021)Dai, Chen, Yang, Zhang, Yuan, and Zhang}]{dai2021dynamic}
\bibinfo{author}{X.~Dai}, \bibinfo{author}{Y.~Chen}, \bibinfo{author}{J.~Yang}, \bibinfo{author}{P.~Zhang}, \bibinfo{author}{L.~Yuan}, \bibinfo{author}{L.~Zhang},
\newblock \bibinfo{title}{Dynamic detr: End-to-end object detection with dynamic attention},
\newblock in: \bibinfo{booktitle}{Proceedings of the IEEE/CVF international conference on computer vision}, \bibinfo{year}{2021}, pp. \bibinfo{pages}{2988--2997}.
\bibitem[{Zhu et~al.(2020)Zhu, Su, Lu, Li, Wang, and Dai}]{zhu2020deformable}
\bibinfo{author}{X.~Zhu}, \bibinfo{author}{W.~Su}, \bibinfo{author}{L.~Lu}, \bibinfo{author}{B.~Li}, \bibinfo{author}{X.~Wang}, \bibinfo{author}{J.~Dai},
\newblock \bibinfo{title}{Deformable detr: Deformable transformers for end-to-end object detection},
\newblock \bibinfo{journal}{arXiv preprint arXiv:2010.04159}  (\bibinfo{year}{2020}).
\bibitem[{Hosang et~al.(2017)Hosang, Benenson, and Schiele}]{hosang2017learning}
\bibinfo{author}{J.~Hosang}, \bibinfo{author}{R.~Benenson}, \bibinfo{author}{B.~Schiele},
\newblock \bibinfo{title}{Learning non-maximum suppression},
\newblock in: \bibinfo{booktitle}{Proceedings of the IEEE conference on computer vision and pattern recognition}, \bibinfo{year}{2017}, pp. \bibinfo{pages}{4507--4515}.
\bibitem[{Carion et~al.(2020)Carion, Massa, Synnaeve, Usunier, Kirillov, and Zagoruyko}]{carion2020end}
\bibinfo{author}{N.~Carion}, \bibinfo{author}{F.~Massa}, \bibinfo{author}{G.~Synnaeve}, \bibinfo{author}{N.~Usunier}, \bibinfo{author}{A.~Kirillov}, \bibinfo{author}{S.~Zagoruyko},
\newblock \bibinfo{title}{End-to-end object detection with transformers},
\newblock in: \bibinfo{booktitle}{European conference on computer vision}, \bibinfo{organization}{Springer}, \bibinfo{year}{2020}, pp. \bibinfo{pages}{213--229}.
\bibitem[{Ren and Ramanan(2013)}]{ren2013histograms}
\bibinfo{author}{X.~Ren}, \bibinfo{author}{D.~Ramanan},
\newblock \bibinfo{title}{Histograms of sparse codes for object detection},
\newblock in: \bibinfo{booktitle}{Proceedings of the IEEE conference on computer vision and pattern recognition}, \bibinfo{year}{2013}, pp. \bibinfo{pages}{3246--3253}.
\bibitem[{Xie et~al.(2013)Xie, Zhang, Li, Lin, Qu, and Zhang}]{xie2013discriminative}
\bibinfo{author}{Y.~Xie}, \bibinfo{author}{W.~Zhang}, \bibinfo{author}{C.~Li}, \bibinfo{author}{S.~Lin}, \bibinfo{author}{Y.~Qu}, \bibinfo{author}{Y.~Zhang},
\newblock \bibinfo{title}{Discriminative object tracking via sparse representation and online dictionary learning},
\newblock \bibinfo{journal}{IEEE transactions on cybernetics} \bibinfo{volume}{44} (\bibinfo{year}{2013}) \bibinfo{pages}{539--553}.
\bibitem[{O'shea and Nash(2015)}]{o2015introduction}
\bibinfo{author}{K.~O'shea}, \bibinfo{author}{R.~Nash},
\newblock \bibinfo{title}{An introduction to convolutional neural networks},
\newblock \bibinfo{journal}{arXiv preprint arXiv:1511.08458}  (\bibinfo{year}{2015}).
\bibitem[{Soydaner(2022)}]{soydaner2022attention}
\bibinfo{author}{D.~Soydaner},
\newblock \bibinfo{title}{Attention mechanism in neural networks: where it comes and where it goes},
\newblock \bibinfo{journal}{Neural Computing and Applications} \bibinfo{volume}{34} (\bibinfo{year}{2022}) \bibinfo{pages}{13371--13385}.
\bibitem[{Khan et~al.(2023)Khan, Rauf, Sohail, Khan, Asif, Asif, and Farooq}]{khan2023survey}
\bibinfo{author}{A.~Khan}, \bibinfo{author}{Z.~Rauf}, \bibinfo{author}{A.~Sohail}, \bibinfo{author}{A.~R. Khan}, \bibinfo{author}{H.~Asif}, \bibinfo{author}{A.~Asif}, \bibinfo{author}{U.~Farooq},
\newblock \bibinfo{title}{A survey of the vision transformers and their cnn-transformer based variants},
\newblock \bibinfo{journal}{Artificial Intelligence Review} \bibinfo{volume}{56} (\bibinfo{year}{2023}) \bibinfo{pages}{2917--2970}.
\bibitem[{Alzubaidi et~al.(2021)Alzubaidi, Zhang, Humaidi, Al-Dujaili, Duan, Al-Shamma, Santamar{\'\i}a, Fadhel, Al-Amidie, and Farhan}]{alzubaidi2021review}
\bibinfo{author}{L.~Alzubaidi}, \bibinfo{author}{J.~Zhang}, \bibinfo{author}{A.~J. Humaidi}, \bibinfo{author}{A.~Al-Dujaili}, \bibinfo{author}{Y.~Duan}, \bibinfo{author}{O.~Al-Shamma}, \bibinfo{author}{J.~Santamar{\'\i}a}, \bibinfo{author}{M.~A. Fadhel}, \bibinfo{author}{M.~Al-Amidie}, \bibinfo{author}{L.~Farhan},
\newblock \bibinfo{title}{Review of deep learning: concepts, cnn architectures, challenges, applications, future directions},
\newblock \bibinfo{journal}{Journal of big Data} \bibinfo{volume}{8} (\bibinfo{year}{2021}) \bibinfo{pages}{1--74}.
\bibitem[{Chen et~al.(2018)Chen, Liu, Gao, and Han}]{chen2018mobilefacenets}
\bibinfo{author}{S.~Chen}, \bibinfo{author}{Y.~Liu}, \bibinfo{author}{X.~Gao}, \bibinfo{author}{Z.~Han},
\newblock \bibinfo{title}{Mobilefacenets: Efficient cnns for accurate real-time face verification on mobile devices},
\newblock in: \bibinfo{booktitle}{Chinese conference on biometric recognition}, \bibinfo{organization}{Springer}, \bibinfo{year}{2018}, pp. \bibinfo{pages}{428--438}.
\bibitem[{Girshick et~al.(2014)Girshick, Donahue, Darrell, and Malik}]{girshick2014rich}
\bibinfo{author}{R.~Girshick}, \bibinfo{author}{J.~Donahue}, \bibinfo{author}{T.~Darrell}, \bibinfo{author}{J.~Malik},
\newblock \bibinfo{title}{Rich feature hierarchies for accurate object detection and semantic segmentation},
\newblock in: \bibinfo{booktitle}{Proceedings of the IEEE conference on computer vision and pattern recognition}, \bibinfo{year}{2014}, pp. \bibinfo{pages}{580--587}.
\bibitem[{Sapkota and Karkee(2024)}]{sapkota2024comparing}
\bibinfo{author}{R.~Sapkota}, \bibinfo{author}{M.~Karkee},
\newblock \bibinfo{title}{Comparing yolov11 and yolov8 for instance segmentation of occluded and non-occluded immature green fruits in complex orchard environment},
\newblock \bibinfo{journal}{arXiv preprint arXiv:2410.19869}  (\bibinfo{year}{2024}).
\bibitem[{Tian et~al.(2025)Tian, Ye, and Doermann}]{tian2025yolov12}
\bibinfo{author}{Y.~Tian}, \bibinfo{author}{Q.~Ye}, \bibinfo{author}{D.~Doermann},
\newblock \bibinfo{title}{Yolov12: Attention-centric real-time object detectors},
\newblock \bibinfo{journal}{arXiv preprint arXiv:2502.12524}  (\bibinfo{year}{2025}).
\bibitem[{Sapkota et~al.(2024)Sapkota, Qureshi, Calero, Badjugar, Nepal, Poulose, Zeno, Vaddevolu, Khan, Shoman et~al.}]{sapkota2024yolov10}
\bibinfo{author}{R.~Sapkota}, \bibinfo{author}{R.~Qureshi}, \bibinfo{author}{M.~F. Calero}, \bibinfo{author}{C.~Badjugar}, \bibinfo{author}{U.~Nepal}, \bibinfo{author}{A.~Poulose}, \bibinfo{author}{P.~Zeno}, \bibinfo{author}{U.~B.~P. Vaddevolu}, \bibinfo{author}{S.~Khan}, \bibinfo{author}{M.~Shoman}, et~al.,
\newblock \bibinfo{title}{Yolov10 to its genesis: a decadal and comprehensive review of the you only look once (yolo) series},
\newblock \bibinfo{journal}{arXiv preprint arXiv:2406.19407}  (\bibinfo{year}{2024}).
\bibitem[{Sapkota and Karkee(2025)}]{sapkota2025improved}
\bibinfo{author}{R.~Sapkota}, \bibinfo{author}{M.~Karkee},
\newblock \bibinfo{title}{Improved yolov12 with llm-generated synthetic data for enhanced apple detection and benchmarking against yolov11 and yolov10},
\newblock \bibinfo{journal}{arXiv preprint arXiv:2503.00057}  (\bibinfo{year}{2025}).
\bibitem[{Meng et~al.(2025)Meng, Du, Sapkota, Ma, and Cheng}]{meng2025yolov10}
\bibinfo{author}{Z.~Meng}, \bibinfo{author}{X.~Du}, \bibinfo{author}{R.~Sapkota}, \bibinfo{author}{Z.~Ma}, \bibinfo{author}{H.~Cheng},
\newblock \bibinfo{title}{Yolov10-pose and yolov9-pose: Real-time strawberry stalk pose detection models},
\newblock \bibinfo{journal}{Computers in Industry} \bibinfo{volume}{165} (\bibinfo{year}{2025}) \bibinfo{pages}{104231}.
\bibitem[{Liu et~al.(2016)Liu, Anguelov, Erhan, Szegedy, Reed, Fu, and Berg}]{liu2016ssd}
\bibinfo{author}{W.~Liu}, \bibinfo{author}{D.~Anguelov}, \bibinfo{author}{D.~Erhan}, \bibinfo{author}{C.~Szegedy}, \bibinfo{author}{S.~Reed}, \bibinfo{author}{C.-Y. Fu}, \bibinfo{author}{A.~C. Berg},
\newblock \bibinfo{title}{Ssd: Single shot multibox detector},
\newblock in: \bibinfo{booktitle}{Computer Vision--ECCV 2016: 14th European Conference, Amsterdam, The Netherlands, October 11--14, 2016, Proceedings, Part I 14}, \bibinfo{organization}{Springer}, \bibinfo{year}{2016}, pp. \bibinfo{pages}{21--37}.
\bibitem[{Lin et~al.(2017)Lin, Goyal, Girshick, He, and Doll{\'a}r}]{lin2017focal}
\bibinfo{author}{T.-Y. Lin}, \bibinfo{author}{P.~Goyal}, \bibinfo{author}{R.~Girshick}, \bibinfo{author}{K.~He}, \bibinfo{author}{P.~Doll{\'a}r},
\newblock \bibinfo{title}{Focal loss for dense object detection},
\newblock in: \bibinfo{booktitle}{Proceedings of the IEEE international conference on computer vision}, \bibinfo{year}{2017}, pp. \bibinfo{pages}{2980--2988}.
\bibitem[{Wang et~al.(2022)Wang, Li, Du, Ma, and Liu}]{wang2022farmland}
\bibinfo{author}{D.~Wang}, \bibinfo{author}{Z.~Li}, \bibinfo{author}{X.~Du}, \bibinfo{author}{Z.~Ma}, \bibinfo{author}{X.~Liu},
\newblock \bibinfo{title}{Farmland obstacle detection from the perspective of uavs based on non-local deformable detr},
\newblock \bibinfo{journal}{Agriculture} \bibinfo{volume}{12} (\bibinfo{year}{2022}) \bibinfo{pages}{1983}.
\bibitem[{Lin et~al.(2024)Lin, Liu, Li, Wei, Liu, Han, and Wu}]{lin2024dcea}
\bibinfo{author}{H.~Lin}, \bibinfo{author}{J.~Liu}, \bibinfo{author}{X.~Li}, \bibinfo{author}{L.~Wei}, \bibinfo{author}{Y.~Liu}, \bibinfo{author}{B.~Han}, \bibinfo{author}{Z.~Wu},
\newblock \bibinfo{title}{Dcea: Detr with concentrated deformable attention for end-to-end ship detection in sar images},
\newblock \bibinfo{journal}{IEEE Journal of Selected Topics in Applied Earth Observations and Remote Sensing}  (\bibinfo{year}{2024}).
\bibitem[{Zong et~al.(2023)Zong, Song, and Liu}]{zong2023detrs}
\bibinfo{author}{Z.~Zong}, \bibinfo{author}{G.~Song}, \bibinfo{author}{Y.~Liu},
\newblock \bibinfo{title}{Detrs with collaborative hybrid assignments training},
\newblock in: \bibinfo{booktitle}{Proceedings of the IEEE/CVF international conference on computer vision}, \bibinfo{year}{2023}, pp. \bibinfo{pages}{6748--6758}.
\bibitem[{Zhang et~al.(2025)Zhang, Wu, Xu, Xie, and Zhang}]{zhang2025improved}
\bibinfo{author}{Y.~Zhang}, \bibinfo{author}{Y.~Wu}, \bibinfo{author}{H.~Xu}, \bibinfo{author}{Y.~Xie}, \bibinfo{author}{Y.~Zhang},
\newblock \bibinfo{title}{Improved co-detr with dropkey and its application to hot work detection},
\newblock \bibinfo{journal}{Concurrency and Computation: Practice and Experience} \bibinfo{volume}{37} (\bibinfo{year}{2025}) \bibinfo{pages}{e70020}.
\bibitem[{Fang et~al.(2021)Fang, Liao, Wang, Fang, Qi, Wu, Niu, and Liu}]{fang2021you}
\bibinfo{author}{Y.~Fang}, \bibinfo{author}{B.~Liao}, \bibinfo{author}{X.~Wang}, \bibinfo{author}{J.~Fang}, \bibinfo{author}{J.~Qi}, \bibinfo{author}{R.~Wu}, \bibinfo{author}{J.~Niu}, \bibinfo{author}{W.~Liu},
\newblock \bibinfo{title}{You only look at one sequence: Rethinking transformer in vision through object detection},
\newblock \bibinfo{journal}{Advances in Neural Information Processing Systems} \bibinfo{volume}{34} (\bibinfo{year}{2021}) \bibinfo{pages}{26183--26197}.
\bibitem[{Zhao et~al.(2024)Zhao, Lv, Xu, Wei, Wang, Dang, Liu, and Chen}]{zhao2024detrs}
\bibinfo{author}{Y.~Zhao}, \bibinfo{author}{W.~Lv}, \bibinfo{author}{S.~Xu}, \bibinfo{author}{J.~Wei}, \bibinfo{author}{G.~Wang}, \bibinfo{author}{Q.~Dang}, \bibinfo{author}{Y.~Liu}, \bibinfo{author}{J.~Chen},
\newblock \bibinfo{title}{Detrs beat yolos on real-time object detection},
\newblock in: \bibinfo{booktitle}{Proceedings of the IEEE/CVF conference on computer vision and pattern recognition}, \bibinfo{year}{2024}, pp. \bibinfo{pages}{16965--16974}.
\bibitem[{Minderer et~al.(2022)Minderer, Gritsenko, Stone, Neumann, Weissenborn, Dosovitskiy, Mahendran, Arnab, Dehghani, Shen et~al.}]{minderer2022simple}
\bibinfo{author}{M.~Minderer}, \bibinfo{author}{A.~Gritsenko}, \bibinfo{author}{A.~Stone}, \bibinfo{author}{M.~Neumann}, \bibinfo{author}{D.~Weissenborn}, \bibinfo{author}{A.~Dosovitskiy}, \bibinfo{author}{A.~Mahendran}, \bibinfo{author}{A.~Arnab}, \bibinfo{author}{M.~Dehghani}, \bibinfo{author}{Z.~Shen}, et~al.,
\newblock \bibinfo{title}{Simple open-vocabulary object detection},
\newblock in: \bibinfo{booktitle}{European conference on computer vision}, \bibinfo{organization}{Springer}, \bibinfo{year}{2022}, pp. \bibinfo{pages}{728--755}.
\bibitem[{Heigold et~al.(2023)Heigold, Minderer, Gritsenko, Bewley, Keysers, Lu{\v{c}}i{\'c}, Yu, and Kipf}]{heigold2023video}
\bibinfo{author}{G.~Heigold}, \bibinfo{author}{M.~Minderer}, \bibinfo{author}{A.~Gritsenko}, \bibinfo{author}{A.~Bewley}, \bibinfo{author}{D.~Keysers}, \bibinfo{author}{M.~Lu{\v{c}}i{\'c}}, \bibinfo{author}{F.~Yu}, \bibinfo{author}{T.~Kipf},
\newblock \bibinfo{title}{Video owl-vit: Temporally-consistent open-world localization in video},
\newblock in: \bibinfo{booktitle}{Proceedings of the IEEE/CVF International Conference on Computer Vision}, \bibinfo{year}{2023}, pp. \bibinfo{pages}{13802--13811}.
\bibitem[{Wang et~al.(2025)Wang, Huang, Li, Yan, Zhang, Lu, and He}]{wang2025effowt}
\bibinfo{author}{B.~Wang}, \bibinfo{author}{K.~Huang}, \bibinfo{author}{B.~Li}, \bibinfo{author}{Y.~Yan}, \bibinfo{author}{L.~Zhang}, \bibinfo{author}{H.~Lu}, \bibinfo{author}{Y.~He},
\newblock \bibinfo{title}{Effowt: Transfer visual language models to open-world tracking efficiently and effectively},
\newblock \bibinfo{journal}{arXiv preprint arXiv:2504.05141}  (\bibinfo{year}{2025}).
\bibitem[{Zhang et~al.(2022)Zhang, Li, Liu, Zhang, Su, Zhu, Ni, and Shum}]{zhang2022dino}
\bibinfo{author}{H.~Zhang}, \bibinfo{author}{F.~Li}, \bibinfo{author}{S.~Liu}, \bibinfo{author}{L.~Zhang}, \bibinfo{author}{H.~Su}, \bibinfo{author}{J.~Zhu}, \bibinfo{author}{L.~M. Ni}, \bibinfo{author}{H.-Y. Shum},
\newblock \bibinfo{title}{Dino: Detr with improved denoising anchor boxes for end-to-end object detection},
\newblock \bibinfo{journal}{arXiv preprint arXiv:2203.03605}  (\bibinfo{year}{2022}).
\bibitem[{Robicheaux et~al.(2025)Robicheaux, Popov, Madan, Robinson, Nelson, Ramanan, and Peri}]{robicheauxroboflow100}
\bibinfo{author}{P.~Robicheaux}, \bibinfo{author}{M.~Popov}, \bibinfo{author}{A.~Madan}, \bibinfo{author}{I.~Robinson}, \bibinfo{author}{J.~Nelson}, \bibinfo{author}{D.~Ramanan}, \bibinfo{author}{N.~Peri},
\newblock \bibinfo{title}{Roboflow100-vl: A multi-domain object detection benchmark for vision-language models},
\newblock \bibinfo{journal}{Roboflow}  (\bibinfo{year}{2025}).
\bibitem[{Sapkota et~al.(2024{\natexlab{a}})Sapkota, Paudel, and Karkee}]{sapkota2024zero}
\bibinfo{author}{R.~Sapkota}, \bibinfo{author}{A.~Paudel}, \bibinfo{author}{M.~Karkee},
\newblock \bibinfo{title}{Zero-shot automatic annotation and instance segmentation using llm-generated datasets: Eliminating field imaging and manual annotation for deep learning model development},
\newblock \bibinfo{journal}{arXiv preprint arXiv:2411.11285}  (\bibinfo{year}{2024}{\natexlab{a}}).
\bibitem[{Sapkota et~al.(2024{\natexlab{b}})Sapkota, Ahmed, Churuvija, and Karkee}]{sapkota2024immature}
\bibinfo{author}{R.~Sapkota}, \bibinfo{author}{D.~Ahmed}, \bibinfo{author}{M.~Churuvija}, \bibinfo{author}{M.~Karkee},
\newblock \bibinfo{title}{Immature green apple detection and sizing in commercial orchards using yolov8 and shape fitting techniques},
\newblock \bibinfo{journal}{IEEE Access} \bibinfo{volume}{12} (\bibinfo{year}{2024}{\natexlab{b}}) \bibinfo{pages}{43436--43452}.
\bibitem[{Sapkota and Karkee(2024)}]{sapkota2024yolo11}
\bibinfo{author}{R.~Sapkota}, \bibinfo{author}{M.~Karkee},
\newblock \bibinfo{title}{Yolo11 and vision transformers based 3d pose estimation of immature green fruits in commercial apple orchards for robotic thinning},
\newblock \bibinfo{journal}{arXiv preprint arXiv:2410.19846}  (\bibinfo{year}{2024}).
\bibitem[{Liu et~al.(2024)Liu, Meng, Zhao, Ma, Zhang, and Jia}]{liu2024green}
\bibinfo{author}{Q.~Liu}, \bibinfo{author}{H.~Meng}, \bibinfo{author}{R.~Zhao}, \bibinfo{author}{X.~Ma}, \bibinfo{author}{T.~Zhang}, \bibinfo{author}{W.~Jia},
\newblock \bibinfo{title}{Green apple detector based on optimized deformable detection transformer},
\newblock \bibinfo{journal}{Agriculture} \bibinfo{volume}{15} (\bibinfo{year}{2024}) \bibinfo{pages}{75}.
\bibitem[{Sapkota et~al.(2025)Sapkota, Raza, Shoman, Paudel, and Karkee}]{sapkota2025multimodal}
\bibinfo{author}{R.~Sapkota}, \bibinfo{author}{S.~Raza}, \bibinfo{author}{M.~Shoman}, \bibinfo{author}{A.~Paudel}, \bibinfo{author}{M.~Karkee},
\newblock \bibinfo{title}{Multimodal large language models for image, text, and speech data augmentation: A survey},
\newblock \bibinfo{journal}{arXiv preprint arXiv:2501.18648}  (\bibinfo{year}{2025}).
\bibitem[{Liu et~al.(2024)Liu, Lv, and Zhang}]{liu2024mae}
\bibinfo{author}{Q.~Liu}, \bibinfo{author}{J.~Lv}, \bibinfo{author}{C.~Zhang},
\newblock \bibinfo{title}{Mae-yolov8-based small object detection of green crisp plum in real complex orchard environments},
\newblock \bibinfo{journal}{Computers and Electronics in Agriculture} \bibinfo{volume}{226} (\bibinfo{year}{2024}) \bibinfo{pages}{109458}.
\bibitem[{Lv et~al.(2024)Lv, Wu, Zhou, Huang, Liu, Gu, Rong, and Zou}]{lv2024fcae}
\bibinfo{author}{J.~Lv}, \bibinfo{author}{Z.~Wu}, \bibinfo{author}{P.~Zhou}, \bibinfo{author}{J.~Huang}, \bibinfo{author}{G.~Liu}, \bibinfo{author}{Y.~Gu}, \bibinfo{author}{H.~Rong}, \bibinfo{author}{L.~Zou},
\newblock \bibinfo{title}{Fcae-yolov8n: a target detection method for immature grape clusters},
\newblock \bibinfo{journal}{New Zealand Journal of Crop and Horticultural Science}  (\bibinfo{year}{2024}) \bibinfo{pages}{1--19}.

\end{thebibliography}

\vspace{2 cm}


\end{document}